\documentclass[runningheads]{llncs}
\usepackage{comment}
\usepackage[pdftex]{graphicx}
\usepackage{caption}
\usepackage{subcaption}
\usepackage{color}
\usepackage{xcolor}
\usepackage{amsmath,amssymb}
\usepackage[ruled, vlined, linesnumbered]{algorithm2e}
\usepackage{multirow}
\usepackage{array}
\usepackage{units}
\pdfoutput=1
\graphicspath{{figures/}}
\DeclareGraphicsExtensions{.pdf,.jpeg,.png,.jpg}

\usepackage[pagebackref=true,breaklinks=true,letterpaper=true,colorlinks,bookmarks=false]{hyperref}

\usepackage{xspace}
\def\sota{state-of-the-art}

\newcommand{\dbase}{\mathcal{D}_b}
\newcommand{\dnovel}{\mathcal{D}_n}
\newcommand{\Spred}{\bar{S}}
\newcommand{\Sprior}{S^p}
\newcommand{\D}{D}

\DeclareMathOperator*{\argmin}{arg\,min}

\usepackage[width=122mm,left=12mm,paperwidth=146mm,height=193mm,top=12mm,paperheight=217mm]{geometry}

\usepackage[font=small,labelfont=bf,tableposition=top]{caption}

\begin{document}
\pagestyle{headings}
\mainmatter
\def\ECCVSubNumber{5140}  

\title{Few-Shot Single-View 3-D Object Reconstruction with Compositional Priors} 


\author{
    Mateusz Michalkiewicz\inst{1} \and
    Sarah Parisot\inst{2,5}\textsuperscript{*} \and
    Stavros Tsogkas\inst{3,4}\thanks{\scriptsize{Stavros Tsogkas and Sarah Parisot contributed to this article in their personal capacity as an
Adjunct Professor at the University of Toronto and Visiting Scholar at Mila, respectively. The views expressed (or the conclusions reached) are their own and do not necessarily represent the views of Samsung Research America, Inc and Huawei Technologies Co., Ltd.}} \and
    Mahsa Baktashmotlagh\inst{1} \and
    Anders Eriksson\inst{1} \and 
    Eugene Belilovsky\inst{2}
}
\authorrunning{Michalkiewicz et al.}
%
\institute{
    University of Queensland \email{m.michalkiewicz@uq.net.au}, \\
    \email{\{m.baktashmotlagh, a.eriksson\}@uq.edu.au} \and 
    Mila, University of Montreal \email{eugene.belilovsky@umontreal.ca}  \and 
    University of Toronto \email{tsogkas@cs.toronto.edu} \and
    Samsung AI Research Center, Toronto \and
    Huawei Noah's Ark Lab. London \email{sarah.parisot@huawei.com}
}
\maketitle

\begin{abstract}
The impressive performance of deep convolutional neural networks in single-view 3D reconstruction
suggests that these models perform non-trivial reasoning about the 3D structure of the output space.
However, recent work has challenged this belief, showing that complex encoder-decoder architectures 
perform similarly to nearest-neighbor baselines or simple linear decoder models that exploit large amounts of per category data in standard benchmarks. 
A more realistic setting, however, involves inferring the 3D shape of objects with few available examples; 
this requires a model that can successfully \emph{generalize} to novel object classes.
In this work we demonstrate experimentally that naive baselines fail in this \emph{few-shot} learning setting, 
where the network must learn informative shape priors for inference of new categories.
We propose three ways to learn a class-specific global shape prior, directly from data.
Using these techniques, our learned prior is able to
capture multi-scale information about the 3D shape, account for intra-class variability by virtue of 
an implicit compositional structure. 
Experiments on the popular ShapeNet dataset show that our method outperforms a zero-shot baseline
by over $50\%$ and the current \sota\ by over $10\%$ in terms of relative performance, 
in the few-shot setting.

\keywords{3D reconstruction, few-shot learning, compositionality,  CNN}
\end{abstract}
\section{Introduction} \label{sec:introduction}
\begin{figure}[t]
    \centering
    \def\w{1}
    \includegraphics[width=\w\textwidth]{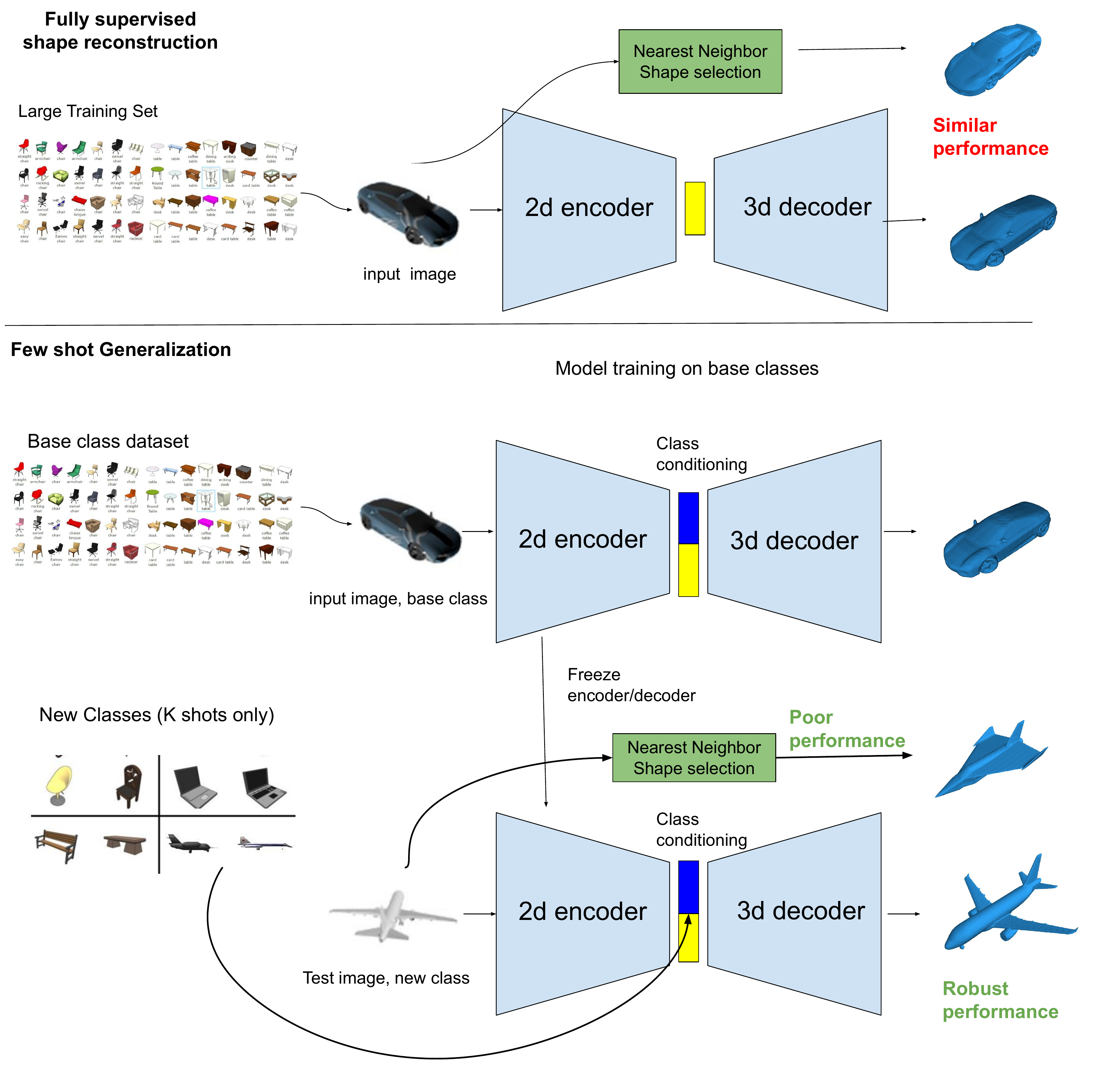}
    \caption{Overview illustrating the differences between the standard 3D object reconstruction and the few-shot generalization setting. The former can be addressed by simple baselines not requiring generalization about shape.}
    \label{fig:overview}
\end{figure}
Inferring the 3D geometry of an object, or a scene, from its 2D projection on the image plane is 
a classical computer vision problem with a plethora of applications, including
object recognition, scene understanding, medical diagnosis, animation, and more.
After decades of research this problem remains challenging as it is
inherently ill-posed: there are many valid 3D objects shapes that correspond to the same 2D
projection.

Traditional multi-view geometry and shape-from-X methods try to resolve this ambiguity
by using multiple images of the same object/scene from different viewpoints to find a 
mathematical solution to the inverse 2D-to-3D reconstruction mapping.
Notable examples of such methods include \cite{savarese20073d,kutulakos2000theory,witkin1981recovering,horn,hoiem2005automatic}.

In contrast to the challenges faced by all these methods, 
humans can solve this ill-posed problem relatively easily, even using \emph{just a single image}.
Through experience and interaction with objects, people accumulate prior knowledge about their 
3D structure, and develop mental models of the world, that allow them to accurately 
predict how a 2D scene could be ``lifted'' in 3D, or how an object would look from a different viewpoint.

The question then becomes: ``how can we incorporate similar priors into our models?''.
Early works use CAD models\cite{kong2017using}.
Xu et al.~\cite{xu2014true2form} use low-level priors and mid-level Gestalt principles such as curvature, 
symmetry, and parallelism, to regularize the 3D reconstruction of a 2D sketch.
However, all these methods require an extremely specific specification of the model priors, and do not permit easily learning about shapes from data, which is impractical and does 
not capture the intuitive way people reason about the 3D world. 

Motivated by the success of deep convolutional networks (CNN) in multiple domains, 
the community has recently switched to an alternative paradigm, where more sophisticated 
priors are directly learned from data.
The idea is straightforward:
given a an appropriate set of paired data, one can train a model that takes as input a 2D image and 
outputs a 3D shape.
Most of these works rely on an encoder-decoder architecture, where the encoder extracts a latent 
representation of the object depicted in the image, and the decoder maps that representation into
a 3D shape \cite{Matryoshka,choy20163d,mescheder2018occupancy}.
Many works have studied ways to make the 3-D decoder more efficient, attempting to improve the shape representation.
Looking at the high quality outputs obtained, it is reasonable to assume that, indeed, 
these models learn to perform non-trivial reasoning about 3D object structure.

Surprisingly, recent works \cite{michalkiewicz2020a,tatarchenko2019single} have shown that 
this is not the case.
Tatar\-chenko et al.~\cite{tatarchenko2019single} argue that, because of the way current benchmarks
are constructed, even the most sophisticated learning methods end up finding shortcuts, 
and can rely primarily on recognition to address the single-view 3D reconstruction task.
Their experiments show that modern CNNs for 3D reconstruction are outperformed by simple
nearest neighbor (NN) or classification baselines, both quantitatively, and qualitatively. Similarly \cite{michalkiewicz2020a} showed that simple linear decoder models learned by PCA are sufficient to achieve high performance. There is one caveat though: to achieve good performance with these baselines, 
having a large dataset is crucial.
More importantly, true 3D shape understanding implies good \emph{generalization} 
to new object classes.
This is trivial to humans --we can reason about the 3D structure of unknown objects, 
drawing on our inductive bias from \emph{similar} objects we have seen--
but still remains an open computer vision problem. 

Based on this observation, we argue that single-view 3D reconstruction is of particular
interest in the few-shot learning setting. Our hypothesis is that learning to recover 3D shapes using few examples, while promoting
generalization to novel classes, provides a good setup for the 
development and evaluation of models that go beyond simple categorization and actually learn about shape. 

To the best of our knowledge, the first work of that kind is by Wallace and Hariharan~\cite{wallace2019few}.
Instead of directly learning a mapping from 2D images to 3D shapes, \cite{wallace2019few}
train a model that uses features extracted from  2D images to refine an input \emph{shape prior} into a final output.
Their framework allows one to easily adapt the shape prior and use it when inferring new classes. 
However their approach has several restrictions. Firstly, when presented with multiple examples of a new class shape the shapes are averaged, or alternatively a random one is selected, both approaches leading to a collapse of intra-class variability. Secondly, the method does not explicitly force inter-class concepts to be learned. 

In this work we first demonstrate empirically that the few-shot generalization baseline is not susceptible to naive benchmarks described in \cite{tatarchenko2019single}. We then address the aforementioned shortcoming of
\cite{wallace2019few} on this task. We investigate three strategies to construct the shape prior, focusing on modelling intra-class variability, compositionality and multi-scale conditioning. More specifically, we first learn a shape prior that captures intra-class variability  by solving an optimization problem involving all shapes available for the new class. We then introduce a compositional bias in the shape prior that allows to build shared concepts across classes that can be transferred to new ones. Finally, we make use of conditional batchnorm to impose class conditioning explicitly at multiple scales of the decoding process.

In summary, we make the following contributions:
\begin{itemize}
    \item We investigate the few-shot learning setting for 3D shape reconstruction and demonstrate that this set-up constitutes an ideal testbed for the development of methods that reason about shapes. 
    \item  We introduce three strategies for shape prior modelling. We notably introduce a compositional approach that successfully exploits similarities across classes.
    \item Extensive experiments demonstrate that we outperform the state of the art by a significant margin  generalizing to new classes more accurately.
\end{itemize}

\section{Related Work} \label{sec:related}


\subsection{Single-view 3D Reconstruction} \label{sec:related:reconstruction}

Recently a focus in 3D reconstruction from single or multiple view has been on finding better shape representations and alternative decoder models to the typically used 3D discretized set of voxels and corresponding 3D CNN decoder. \cite{choy20163d,girdhar2016learning,wu2016learning,yan2016perspective,Zhu_2017_ICCV}, one that can permit more efficient learning and generation. These include point clouds \cite{fan2017point}, meshes \cite{wang2018pixel2mesh}, signed distance transform based representations \cite{park2019deepsdf,michalkiewicz2020a}. Indeed there is not an agreed upon canonical 3D shape representation and decoder structure for use with deep learning models. However, \cite{tatarchenko2019single,michalkiewicz2020a} has shown that although many of these models improve over each other they do not beat naive baselines such as nearest neighbors.



\subsection{Few-shot Learning} \label{sec:related:fewshot}
Few-shot learning has become a highly popular research topic in computer vision and machine learning \cite{Ravi2017OptimizationAA,gidaris2018dynamic}. Typically research focuses on the classification task, with few works investigating more complex problems such as segmentation \cite{siam2019adaptive} or object detection. Existing methods comprise two categories: meta-learning/meta-gradient based approaches \cite{finn2017model}, and metric-learning \cite{qi2018low,snell2017prototypical}. The former aims to teach models to adapt quickly, in a few gradient updates, to new unseen classes, while the latter learns a distance metric such that the distance of a query image to the few annotated examples of the same class is minimal.  




\section{Methods} \label{sec:methods}
Let $\dbase = \{(I_i, S_i)\}_{i=1}^K$ be a set of $K$ image-shape pairs, belonging to one of $N_b$ \emph{base} object classes.
We assume that $|\dbase|$ is \emph{large}, i.e., $\dbase$  contains enough training examples for our purposes.
We also consider a \emph{much smaller} set $\dnovel$, containing examples of \emph{novel} classes.
Each class in $\dnovel$ possesses only a small set of K image-shape pairs $\{(I^1_n, S^1_n), \ldots, (I^K_n, S^K_n)\}$,
and a large set of test or query images. 
Given $\dbase$, our objective is to use the abundant data in $\dbase$ to 
build a model that takes a 2D image $I$, containing a single object, as input, 
and outputs a 3D reconstruction of the object, $\Spred$.
Similar to previous works employing an encoder-decoder architecture, we choose voxels as our 3D shape representation.
At the same time, we also want our model to be able to leverage the limited data in $\dnovel$, 
to successfully generalize to novel categories.

We propose \emph{three} strategies to achieve this. 
We first introduce our global class conditioning approach (GCE), in Section~\ref{sec:methods:GCE}. 
GCE models the shape prior of a specific object class, as a \emph{learned} global embedding. 
In Section~\ref{sec:methods:CGCE}, we describe a compositional extension of GCE, which aims to learn 
compositional representations across classes and exploit their similarities. 
Finally, in \ref{sec:methods:MCGE} we investigate multi-scale conditioning via 
conditional batch normalization~\cite{perez2018film}.

\subsection{Shape Encoding and Global Class Embedding} \label{sec:methods:GCE}
\begin{figure}
    \centering
    \def\w{0.7}
    \includegraphics[width=\w\textwidth]{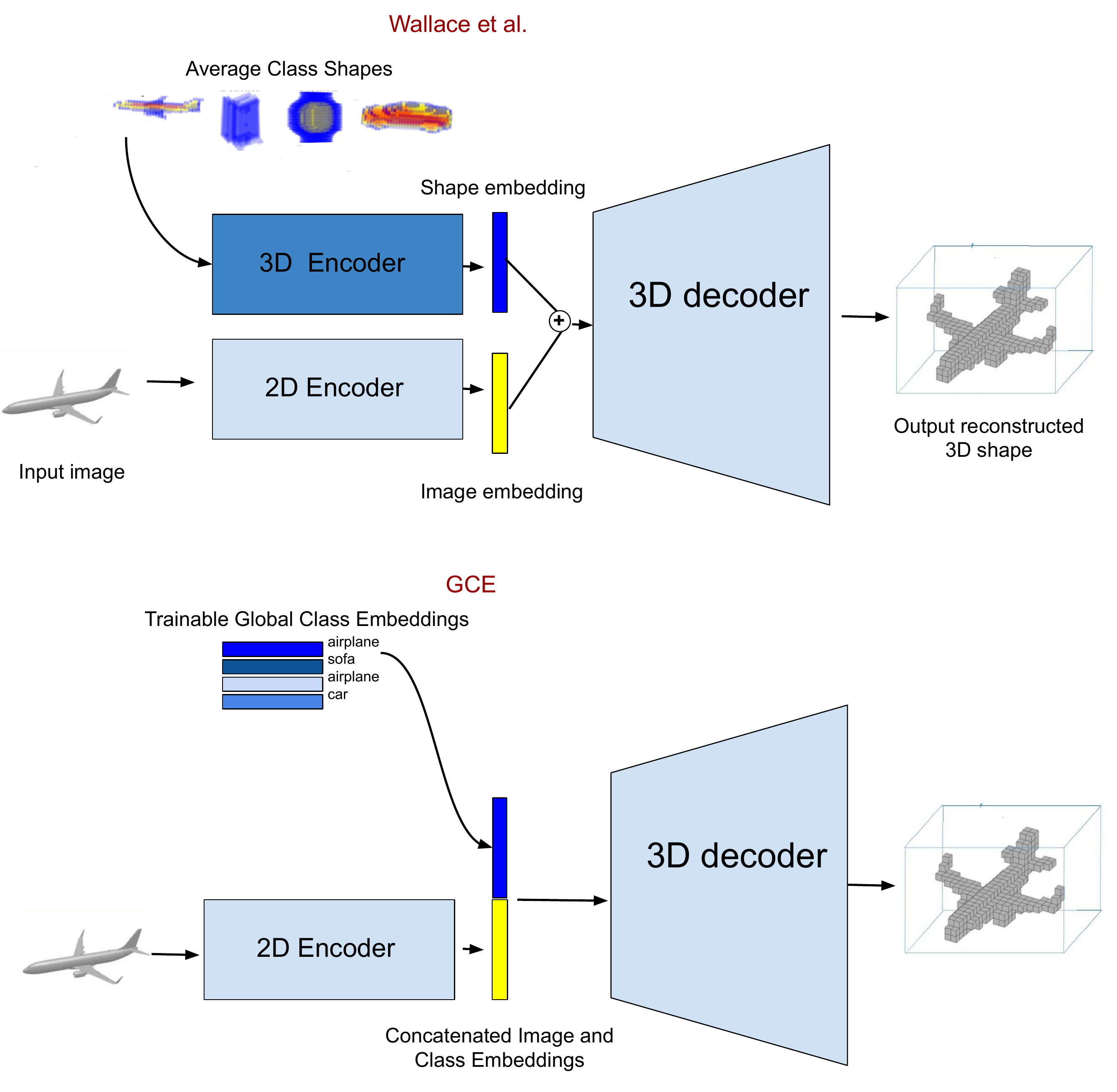}
    \caption{Comparison of \cite{wallace2019few} to GCE. 
    The former collapses variability of new classes by averaging. GCE is able to obtain 
    a global shape representation for each class.}
    \label{fig:GCE}
\end{figure}

Consider an encoder-decoder framework involving
\begin{itemize}
    \item an encoder $E_I$ that takes a 2D image, $I$, and outputs its embedding, $e_I$;
    \item a category specific shape embedding, $e_S$; 
    \item a decoder $D$ that takes the image and shape embeddings and outputs the reconstructed 3D shape, $\Spred$,
    in the form of a voxelized grid.
\end{itemize}
This relationship is formally expressed by
\begin{equation}
    \label{eq:model}
    \Spred = \D \left(e_I, e_S \right) = \D \left( E_I(I), e_S\right).
\end{equation}
This model can be trained using a binary cross entropy loss on the voxel occupancy confidence $p_i$ for voxel $i$ 
in the output grid given
\begin{equation}
    \label{eq:loss}
     \mathcal{L}(S, \Spred) = - \frac{1}{N}\sum_i^N y_i \log(p_i) + (1-y_i) \log(1-p_i).
\end{equation}
In the rest of the text, we drop $S$ for notational simplicity.
For base class training, $E_I,E_S$, and $D$ are learned by minimizing \eqref{eq:loss}. 
For inference on examples of new classes, $e_S$ is computed and fed along with extracted image embedding 
as input to the trained network. 

In~\cite{wallace2019few} (illustrated in Figure~\ref{fig:GCE} (top)), $e_S$ is computed with a shape encoder $E_S$ that takes a category-specific shape prior $\Sprior_c$; 
$\Sprior_c$ is either a randomly selected shape from the set of training shapes associated with class $i$, or the average of all training 
shapes for a class in voxel space.
Providing an explicitly defined single shape or average shape as a shape prior has severe limitations.
Such a prior cannot account for intra-class variability and is therefore intrinsically sub-optimal in settings with more than 
one new examples are considered. 
We propose to address this issue by \emph{learning} a global class embedding (GCE), $e_S^i$, that captures the 
``essence'' of object class-$i$.
The GCE is built using all available shapes for a particular class, we expect a high-dimensional latent representation
to be much more successful in capturing nuances (like intra-class variability) than simple shape averaging. 

Our framework is illustrated in Figure~\ref{fig:GCE} (bottom). 
The encoder $E_I$ and decoder $D$  are trained jointly with the base class embeddings $e_S^i$ 
on the set of base classes, minimizing \eqref{eq:loss}.   
For novel classes with a small training set $\{(I_i,S_i)\}_{i=1}^N$, all model parameters of $E_I$ and $D$ are fixed, and class specific embeddings $e_S^i$ are obtained by solving 
$$\hat{e}_S^i = \argmin_{e_S^i} \sum_{j=1}^N\mathcal{L}(D(E_I(I_j),e_S^i))$$ using all of the available training images. We note that, due to our few-shot set-up, this optimization problem can be solved in few iterations since it only involves a small set of parameters ($e_S^i$) and a small amount of new samples. By construction, this model \textit{does not lose performance on base classes} and can continually add multiple new classes, while at the same time learning \emph{implicit} shape prior representation to guide the decoding process. 
 
Finally we note that we use concatenation  at the first stage of $D$  to combine $e_I$ and $e_S$, instead of the element-wise sum of $e_I$ and $e_S$ used in \cite{wallace2019few}.

\subsection{Compositional Global Class Embeddings} \label{sec:methods:CGCE}
GCE allows us to exploit all available, class specific training shapes to learn a representative shape prior. 
However, the learned global embeddings do not explicitly exploit similarities across different 
classes, which may result in sub-optimal, and potentially redundant representations.

\begin{figure}
    \centering
    \def\w{1}
    \includegraphics[width=\w\textwidth]{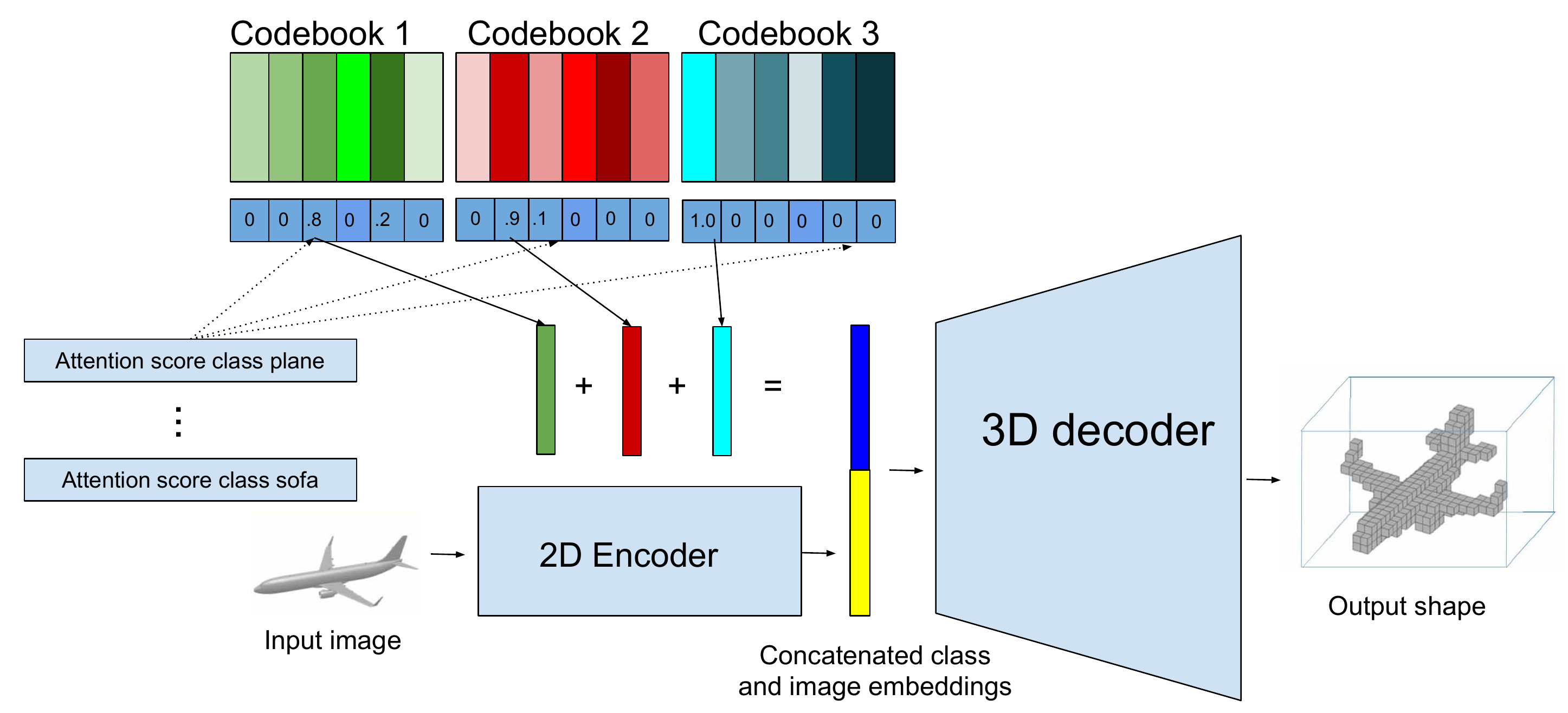}
    \caption{Compositional GCE constructs a code by a composition of codes from different codebooks, applying a different attention to each codebook based on the class. }
    \label{fig:CGCE}
\end{figure}

However, it does not explicitly exploit similarities across shapes, a strategy which would allow increased robustness in the lowest data regimes. We introduce a novel strategy to address this issue, attempting to learn compositional representations between classes which we call Compositional Global Class Embeddings (CGCE). This model is illustrated in Figure~\ref{fig:CGCE}.
 
Our objective is to explicitly encourage the model to discover shared concepts among shapes which can be reused across shapes. 
Taking inspiration from work on compressing word embeddings \cite{shu2017compressing}, we propose to decompose our class representation into a linear combination of vectors that are shared across classes. More specifically, we learn a set of $M$ codebooks (or embedding tables), with each codebook $C_i$ comprising $m$ individual embedding vectors (or \emph{codes}) $C_i = \{e_{i,1},\dots,e_{i,m}\}$, where $e_{i,m} \in \mathcal{R}^D$. Intuitively, each codebook can be interpreted as the representation of an abstract concept which can be shared across multiple classes.

For each class $i$, a learned attention vector $\boldsymbol{\alpha_i}$ selects the most relevant code(s) from each codebook. The codes of all codebooks are then combined to yield an final embedding: $e_S^i = \sum_{k=1}^M \sum_{j=1}^{m}a_i^{k,j} c_{k,j}$ 
where $a_i^{j,k}$ corresponds to scalar attention on the code $j$ at codebook $k$, while $c_{j,k}$ corresponds to the $j^{th}$ code of codebook $k$.  During base class training we learn both $\boldsymbol{\alpha}_i$ and $c_{j,k}$. We highlight that codebooks are shared across classes and therefore need only be trained on base classes. As a result, $\boldsymbol{\alpha}_i$ constitutes the only class specific variable to fine-tune on novel classes: 

 $$\hat{\boldsymbol{\alpha}}_i= \argmin_{\boldsymbol{\alpha}_i} \sum_{j=1}^N\mathcal{L}(D(E_I(I_j),e_S^i)).$$

 Since we would like each codebook to assign a discrete attribute to each class, we aim for the model to select few codes in a given codebook. We thus use a form of attention that relies on the sparsemax \cite{martins2016softmax} operator. Specifically each codebooks attention is given by $a_c^{k} = \textsc{sparsemax}(w_c^k)$, where $w_c^k$ is a learned parameter and the sparsemax operator gives an output that sums to $1$, but will typically attend to just a few outputs. 
 
\subsection{Multi-scale Conditional Class Embeddings} \label{sec:methods:MCGE}
Another inductive bias we can add is the use of multi-scale structure in the shape construction. An elegant way to do this is by applying the conditional batchnorm technique \cite{perez2018film} to the 3D decoder model.  Conditional batch normalization replaces the affine parameters in all batch-norm layers with embeddings at each layer. Since 3D decoders have an inherent multi-scale structure that gradually creates more refined areas in the output, each layer's batch-norm parameters can be seen as defining the class conditional structure at different scales. As in GCE when receiving novel classes we  learn the conditional batchnorm parameters for the new class, keeping all other weights frozen. We refer to this approach as Multi-scale Conditional Class Embeddings (MCCE).

\subsection{Nearest Neighbor Oracle, Zero-Shot and All-Shot Baselines}\label{sec:methods:oracle}
In this section, we introduce and discuss three simple baselines that are used in our experiments.
First, we consider an \emph{oracle nearest neighbor  (ONN)}~\cite{tatarchenko2019single} baseline. 
Given a query 3D shape, ONN exhaustively searches a shape database for the most similar entry with respect to a given metric, 
(Intersection-Over-Union/IoU in this case).
Although this method cannot be applied in practice, it provide an upper bound on how well a retrieval 
method can perform on the task. With ONN, we aim to show that unlike the standard paradigm \cite{tatarchenko2019single}, the few shot generalization benchmark cannot be solved with such a naive baseline.

We further consider a zero-shot (ZS) baseline, and an all-shot (AS) baseline.
For the \emph{ZS baseline}, we train encoder decode model as described in Eq. \eqref{eq:model} and use to infer 3D shapes for novel classes, 
\emph{without} using the category-specific shape prior $e_S$.
We expect this to give a lower bound of performance, since it does not make any use of shape prior information.
For the \emph{AS baseline}, we merge the base class and novel class datasets, obtaining $D_A = D_b \bigcup D_n$.
We train the model on this joint dataset and then test on examples from only the novel classes.
We expect that this baseline will set an upper bound on the performance of the vanilla encoder-decoder architecture, 
since the model also has access to the examples from the novel classes in $D_n$.


\section{Experiments} \label{sec:experiments}

\subsection{Dataset and Evaluation Protocol} \label{sec:experiments:dataset}
For our experiments we use the ShapeNetCore\_v1.0~\cite{chang2015shapenet} dataset and the few-shot generalization benchmark of 
\cite{wallace2019few}.
As in \cite{wallace2019few} we use the following 7 categories as our \emph{base classes}: 
\textbf{plane, car, chair, display, phone, speaker, table}. 
We also use following 10 categories as our \emph{novel classes}: 
\textbf{bench, cabinet, lamp, rifle, sofa, watercraft, knife, bathtub, guitar, laptop}. 
Note that we have added additional categories to the standard benchmark, for a more extensive evaluation. 
Out data comes in the form of pairs of $128\times128$ sRGB images rendered using Blender~\cite{bllender}, 
and $32\times32\times32$ voxelized representations obtained using Binvox~\cite{nooruddin03,binvox}.
Each 3D model has 24 associated images, rendered from random viewpoints.
For evaluation we use the standard Intersection over Union (IoU) score to compare predicted shapes $\Spred$
to ground truth shapes $S$: $\text{IoU} = \vert S \cap \Spred \vert / {\vert S \cup \tilde S \vert}$.

\subsection{Implementation Details} \label{sec:experiments:implementation}
All methods are trained on the 7 base classes except for the AS-baseline which is trained on all 17 categories. 
All methods share the same 2D encoder and 3D decoder architectures.
We use the same 2D encoder as in \cite{Matryoshka,wallace2019few}, a ResNet~\cite{he2015deep} that takes a 
$128\times128$ image as input, and outputs a 128-dimensional embedding.
Our 3D decoder consists of 7 convolutional layers, followed by batch-normalization, 
and ReLU activations. 
For training, we use the same 80-20 train-test split as in R2N2~\cite{choy20163d,wallace2019few}.
Unless otherwise stated, we use $l_r = 0.0001$ as the learning rate and ADAM~\cite{kingma2014adam} as the optimizer. 
All networks are trained with binary cross entropy on the predicted voxel presence probabilities 
in the output 3D grid.

\paragraph{\bf{ZS-Baseline}} is trained on the 7 base categories for 25 epochs. 
We use the trained model to make predictions for novel classes without further adaptations.

\paragraph{\bf{AS-baseline}} is trained on \emph{all} 17 categories for 25 epochs.
Note that we do not use any pre-trained weights, but we rather train this baseline model from a random initialization.

\paragraph{\bf{Wallace et al.~\cite{wallace2019few}.}} To ensure a fair comparison in our experiments, we re-implemented 
this framework, using the exact same settings reported in the respective paper.
In the supplementary material we include a comparison only on the subset of classes used 
in \cite{wallace2019few}, validating that our implementation yields practically identical results.  

\paragraph{\bf{GCE.}} We use the same architecture as in the baselines models and in \cite{wallace2019few}. 
As mentioned in Section~\ref{sec:methods:GCE}, we concatenate the 128-$d$ embeddings from the 2D encoder and 
the conditional branch (as opposed to~\cite{wallace2019few} which uses element-wise addition); 
and we feed the resulting 256-$d$ embedding into the 3D decoder. 
The class conditioning vectors are initialized randomly following a normal distribution $\sim N(0,1)$. 
After training the GCE on the base classes, we freeze the parameters of $E_I$ and $\D$ and
initialize the \emph{novel class} embeddings $c_i$ as the average of the learned base class encodings. 
We then optimize $c_i$ using stochastic gradient descent (SGD) with momentum set to $0.9$.

\paragraph{\bf{CGCE.}} The conditional branch is composed of 5 codebooks, each containing 6 codes of dimension 128,
and an attention array of size $17\times5\times6$; i.e., one attention value per $(class, codebook, code)$ triplet.
The codes and attention values are initialized using a uniform distribution $U(-0.4, 0.4)$. 
During training, we push the attention array to focus on meaningful codes by employing \emph{sparsemax} \cite{martins2016softmax}, 
a modification of the standard softmax function that promotes sparsity of the output.
After training the CGCE on the base classes, we freeze the parameters of $E_I$ and $\D$, as well as the codebook entries $c_{k,j}$. 
We initialize the \emph{novel class} attentions $\boldsymbol{\alpha}_i$ from a uniform distribution $U(-0.4,0.4)$. 
We then optimize $\boldsymbol{\alpha}_i$ using stochastic gradient descent (SGD) with momentum set to $0.9$.

\paragraph{\bf{MCCE}} We replace all batch normalization (bnorm) layers in the 3D decoder with \emph{conditional} 
batch normalization (cond-bnorm)~\cite{perez2018film}. 
More precisely, the affine parameters $\gamma_i$ and $\beta_i$ are initialized from a normal distribution $\sim N(1, 0.2)$,
and conditioned on the class ($i$).
For novel class adaptation only the aforementioned $\gamma_i$ and $\beta_i$ for new classes are learned. 
We use SGD as optimizer with momentum set to 0.9 for this novel class adaptation.
\subsection{Evaluating the Zero-shot Baseline and nearest neighbor oracle} \label{sec:experiments:baselines}

\begin{table}[h]
    \centering
    \resizebox{\textwidth}{!}{
    \begin{tabular}{|c|c|c|c|c|c|c|c|c|c|c|}
        \hline
        cat & ZS-Baseline & AS-baseline & ONN\_1 & ONN\_2 & ONN\_3   & ONN\_4 & ONN\_5 & ONN\_10 & ONN\_25 & ONN\_full\\
        \hline
        bench      & 0.366 & 0.524  & 0.238  & 0.240  &0.245& 0.271      & 0.276  & 0.360 &0.420&0.708 \\
        cabinet    & 0.686 & 0.753  & 0.400  & 0.458  &0.460& 0.461      & 0.480  & 0.495 &0.631&0.842 \\
        lamp       & 0.186 & 0.368  & 0.153  & 0.162  &0.177& 0.189      & 0.194  & 0.223 &0.282&0.515 \\
        firearm    & 0.133 & 0.561  & 0.377  & 0.396  &0.420& 0.425      & 0.434  & 0.510 &0.550&0.707 \\
        sofa     & 0.519 & 0.692  & 0.445  & 0.458  &0.459& 0.530      & 0.534  & 0.579 &0.616&0.791 \\
        watercraft & 0.283 & 0.560  & 0.259  & 0.286  &0.317& 0.354      & 0.372  & 0.479 &0.527&0.697\\
        \hline
        \hline
        mean\_novel &0.362 & 0.576   & 0.312 & 0.333  &0.346& 0.371      & 0.381  & 0.441 &0.504&0.710\\
        \hline
    \end{tabular}
    }

    \caption{Zero-shot, All-shot , and Oracle Nearest Neighbor results for different number of shots. We observe that ONN outperforms the encoder-decoder model on the full dataset while with few samples even the zero-shot approach can outperform the naive oracle nearest neighbors approaches and illustrate generalization.}
    \label{table:onn}
\end{table}
We first illustrate that naive baselines perform poorly in a few-shot setting \cite{wallace2019few} which requires good generalization about shapes. We do this by performing a comparison to two baseline methods shown in Table~\ref{table:onn}. The first is a zero-shot (ZS) baseline, where a model is trained on base classes and used directly to infer shapes of novel classes.  The all-shot (AS) baseline is trained on all classes using the full data (which has 1000-8000 shapes per class). We can expect these two baselines to be upper and lower bounds on the performance of the few-shot generalization benchmark. We use the nearest neighbor oracle described in Sec.~\ref{sec:methods:oracle} for several possible levels. We observe that the full nearest neighbor oracle outperforms the deep learning model trained on all the data (as observed in \cite{wallace2019few}) indicating the task does not require generalization. On the other hand in the few shot case we see that performance is often below that of even the ZS baseline, indicating this task is appropriate for evaluating model generalization.

\begin{figure}
    \centering
    \includegraphics[width=\textwidth]{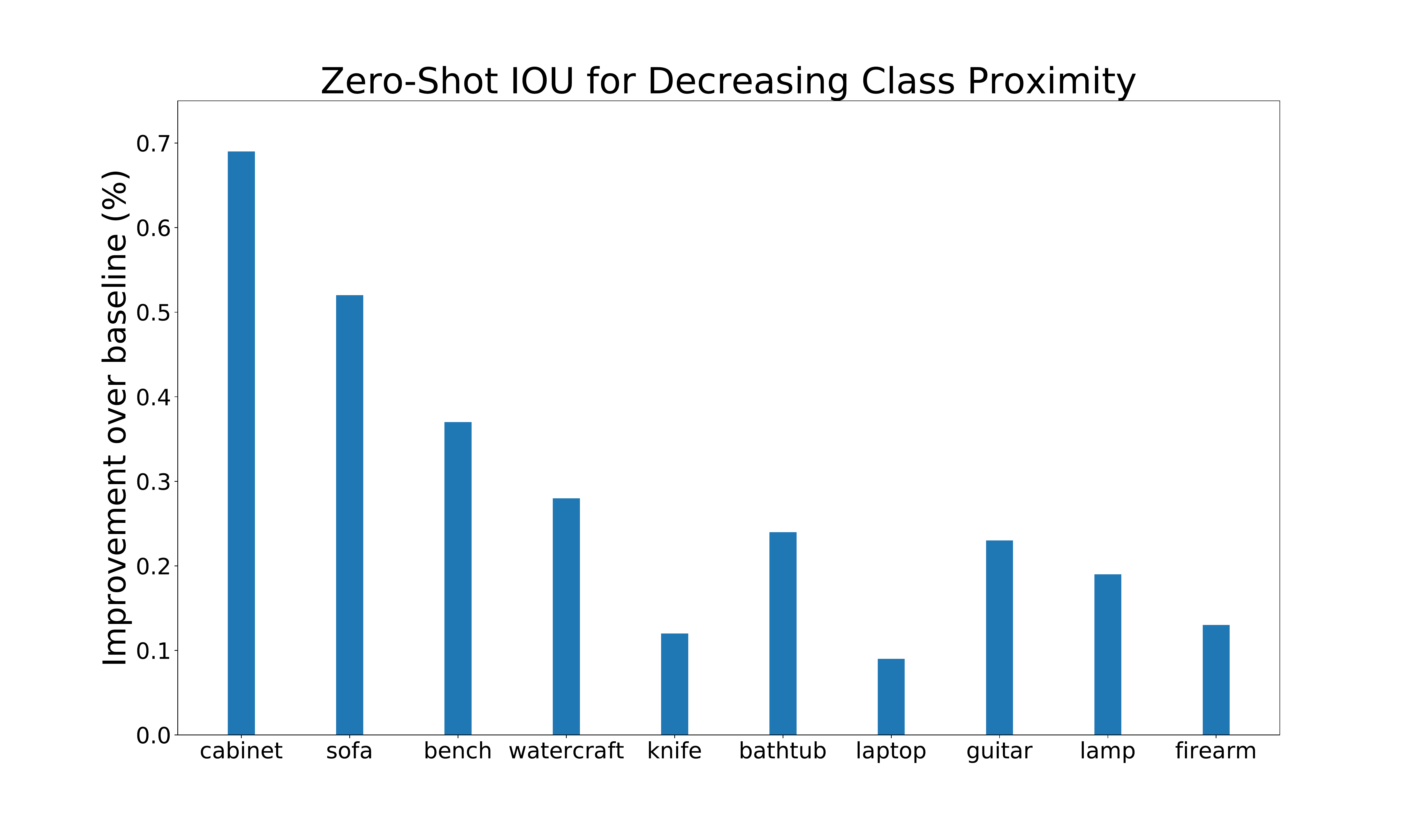}
    \caption{Zero-shot IOU for decreasing proximity}
    \label{fig:prox}
\end{figure}

Note that the ZS benchmark in Table~\ref{table:onn} already obtains relatively high performance on select classes (sofa and cabinet), we hypothesize this is due to similarity towards base classes. In order to best evaluate these methods we take the novel classes and attempt to compute a similarity metric of new classes towards base classes. We do this by finding the average IOU of the nearest neighbor from base classes. Specifically for each shape in the novel class we compute $max_{j \in BaseShapes} IOU(S_{novel}^i,S_{base}^j)$ for all novel shapes $i$ in a given class and average across the class to obtain the proximity scores. The proximity scores for different classes are illustrated in Table~\ref{table:proximity}. We observe in Figure~\ref{fig:prox} that indeed the higher performing zero-shot cases correspond to close proximity classes, we thus use the proximity criteria as an additional consideration when comparing methods. Furthermore to extend the number of distant classes to the base class set, we select 4 additional far proximity classes from  ShapeNet than the original benchmark used in \cite{wallace2019few}, these are included in green in Table~\ref{table:proximity}.        

\subsection{Evaluating Few Shot-Generalization}
\begin{table}
    \centering
    \begin{tabular}{|c|c|}
        \hline

        \hline

        \hline
                        & inter class proximity   \\
        \hline
            base        &  1         \\
        \hline
            cabinet     &   0.794   \\
        \hline
            sofa        &  0.671     \\
        \hline
            bench       &   0.574     \\
        \hline
            watercraft  &  0.539      \\
        \hline
            \textcolor{green}{knife}       &  \textcolor{green}{0.458}      \\
        \hline
            \textcolor{green}{bathtub}     &  \textcolor{green}{0.436}      \\
        \hline
          \textcolor{green}{ laptop}      & \textcolor{green}{0.432}       \\
        \hline
           \textcolor{green}{ guitar}      &  \textcolor{green}{0.429}      \\
        \hline
            lamp        &   0.347     \\
        \hline
            firearm     &   0.290    \\
        \hline
    \end{tabular}   
    \caption{ Proximity between \textit{novel} classes and all base examples. Higher is closer. 
        New classes highlighted
    }
    \label{table:proximity}
\end{table}

We now evaluate the methods discussed in Sec~\ref{sec:methods}.  We evaluate the three methods on the 1-shot baseline in Table~\ref{table:1_shot}. We report both the IOU as well as the relative improvement over the ZS baseline. Note that as the ZS-baseline provides strong performance for easy classes, the average IOU is dominated by these, thus relative improvement is a more meaningful metric for aggregation across classes. 
Observe that GCE improves performance over the method of \cite{wallace2019few}, particularly in cases where the classes are distant, obtaining $45\%$ improvement over Zero-Shot overall compared to \cite{wallace2019few}. CGCE and MCCE give a more drastic improvement in the performance obtaining $54\%$ and $52\%$ respectively by adding multiscale and compositional priors to the simple class prior of \cite{wallace2019few}.

\begin{table}[h]
    \centering
    \resizebox{\textwidth}{!}{
	\begin{tabular}{|c|c|c|c|c|c|c|c|} 
    \hline
    &  0-shot    & All-shot          &    \multicolumn{4}{|c|}{1 shot}   \\
    \hline			 
	cat         & B0     & AS          &       Wallace     & GCE             &  CGCE            & MCCE                          \\

	\hline

		cabinet     & 0.69  & 0.75       & 0.69 (0.00)   & 0.69 (0.01)   & \textbf{0.71 (0.03)}    &  0.69 (0.01)   \\
		sofa        & 0.52  & 0.69       & 0.54 (0.04)   & 0.52 (0.00)  & \textbf{0.54 (0.04)}    &  \textbf{0.54 (0.03)}  \\
		bench       & 0.37  & 0.52       & 0.37 (0.00)   & 0.37 (0.00)   & \textbf{0.37 (0.00)}    &  \textbf{0.37 (0.00)}   \\
		watercraft  & 0.28  & 0.56       & 0.33 (0.16)   & 0.34 (0.19)   & \textbf{0.39 (0.39)}    &  0.37 (0.29)         \\        
		knife       & 0.12  & 0.60       & 0.30 (1.47)   & 0.26 (1.13)   & \textbf{0.31 (1.5)}    &  0.27 (1.19)       \\                               
		bathtub     & 0.24  & 0.46       & 0.26 (0.05)   & 0.27 (0.09)   & \textbf{0.28 (0.13)}    &  0.27 (0.11)               \\                                
		laptop      & 0.09  & 0.56       & 0.21 (1.30)   & 0.27 (1.85)   & \textbf{0.29 (2.10)}    &  0.27 (1.87)       \\                                  
		guitar      & 0.23  & 0.69       & 0.31 (0.38)   & 0.30 (0.31)   & \textbf{0.32 (0.42)}    &  0.30 (0.31)        \\                                
		lamp        & 0.19  & 0.37       & 0.20 (0.05)   & 0.20 (0.07)   & 0.20 (0.05)    &  \textbf{0.22 (0.16)}  \\
		firearm     & 0.13  & 0.56       & 0.21 (0.58)   & 0.24 (0.83)   & 0.23 (0.70)    &  \textbf{0.30 (1.26)}        \\        
		\hline
		mean (relative to B0)  &        &             & 40.2$\%$           & 44.7$\%$           & \textbf{53.7}$\%$            &    52.2$\%$        \\

    \hline
    \end{tabular}
}
	\caption{3D reconstruction from single image in 1-shot setting. Numbers in parenthesis indicate relative performance gain of all novel classes}
\label{table:1_shot}
\end{table}

As discussed in Sec.~\ref{sec:methods} our approach based on global conditional embeddings is able to capture intra-class variability and thus extends more naturally beyond the 1-shot setting. In Table~\ref{table:multishot} we evaluate the compositional method (which performs best in 1-shot evaluation) on 10 and 25-shot settings and compare to \cite{wallace2019few}. We observe that similar to the 1-shot case most methods do not improve much the performance for close-proximity cases. However, for more distant classes we can see substantially performance improvement (sometimes 200$\%$+ in IOU). In Table~\ref{table:multishot} we observe the increased performance of CGCE versus \cite{wallace2019few} for increasing shots. Indeed for larger number of observations the method is able to obtain larger performance gains.  

\begin{table}[t]
\centering
\resizebox{\textwidth}{!}{
	\begin{tabular}{|c|c|c|c|c|c|c|c|} 
\hline
		         &  0-shot    & all-shot          &    \multicolumn{2}{|c|}{10 shot} &  \multicolumn{2}{|c|}{25 shot}  \\
\hline			 
		cat         & B0     & AS          &       Wallace      & CGCE             &  Wallace          & CGCE                          \\
	\hline

		cabinet     & 0.69  & 0.75       &  0.69 (0.00)  & \textbf{0.71 (0.03)}  &   0.69 (0.01)  & \textbf{0.71 (0.04)} \\
		sofa        & 0.52  & 0.69       &  0.54 (0.04)  & 0.54 (0.04)   &   0.54(0.04)  &  \textbf{0.55 (0.06)}   \\
		bench       & 0.37  & 0.52       &  0.36 (-0.01) & \textbf{0.37 (0.03)}   & 0.36 (-0.01)    &  \textbf{0.38 (0.04)}  \\
		watercraft  & 0.28  & 0.56       &  0.36 (0.26)  & \textbf{0.41 (0.45)}   & 0.37 (0.29)    &  \textbf{0.43 (0.53)}   \\ 
		knife       & 0.12  & 0.60       &  0.31 (1.52)  & \textbf{0.32 (1.62)}   & 0.31 (1.57)    &  \textbf{0.35 (1.87)} \\
		bathtub     & 0.24  & 0.46       &  0.26 (0.05)  & \textbf{0.28 (0.16)}   & 0.26 (0.06)    &  \textbf{0.30 (0.23)}\\ 
		laptop      & 0.09  & 0.56       &  0.24 (1.53)  & \textbf{0.30 (2.24)}   & 0.27 (1.85)    &  \textbf{0.32 (2.45)}\\ 
		guitar      & 0.23  & 0.69       &  0.32 (0.39)  & \textbf{0.33 (0.47)}   & 0.32 (0.42)    &  \textbf{0.37 (0.62)}\\ 
		lamp        & 0.19  & 0.37       &  0.19 (0.04)  & \textbf{0.20 (0.05)}   & 0.19 (0.03)    &  \textbf{0.20 (0.07)}  \\
		firearm     & 0.13  & 0.56       &  \textbf{0.24 (0.83)}  & 0.23 (0.75)   & 0.26 (0.95)    &  \textbf{0.28 (1.08)}  \\ 
		\hline
		mean (relative to B0)  &        &             &  46.5$\%$          & \textbf{58.3}$\%$          &   51.9$\%$          &  \textbf{69.8}$\%$       \\

		\hline
\end{tabular}
}
	\caption{3D reconstruction from single image in 10-shot and 25-shot setting. Numbers in parenthesis indicate performance gain over B0. We observe widening improvement with more examples using CGCE.  }
    \label{table:multishot}
\end{table}

\begin{figure}[ht]
    \centering
    \includegraphics[width=0.8\textwidth]{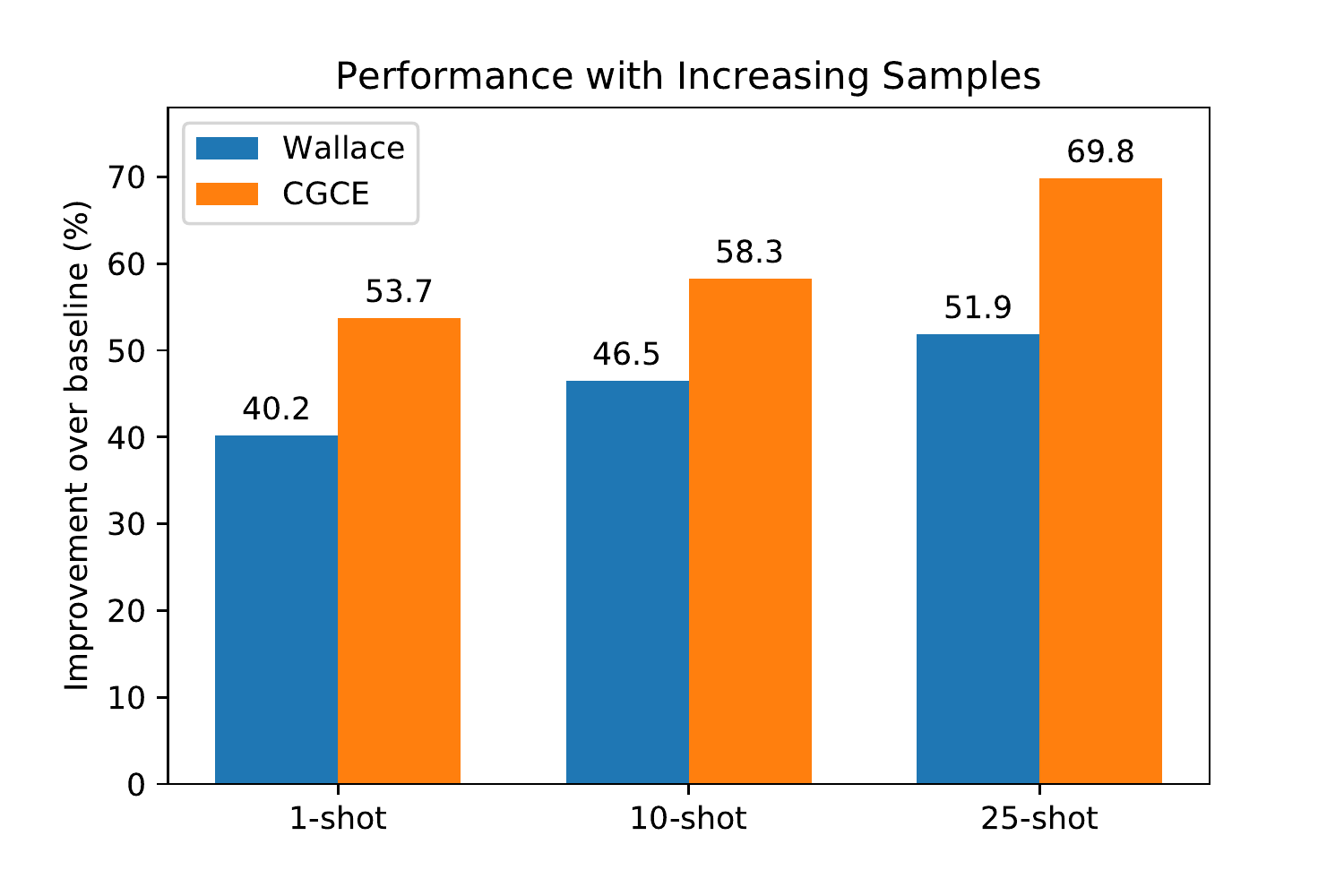}
    \caption{Percentage gains for 1, 10 and 25 shot over B0 baseline. Our method significantly 
        outperforms \cite{wallace2019few} for a larger number of shots. \vspace{10pt}}\label{fig:multishot
    }
\end{figure}

\begin{table}[ht]
    \centering
    \begin{tabular}{|c|c|c|c|c|}
        \hline
        cat                 & B0 & AS &         GCE    & GCE\_rand                       \\
        \hline
        plane               & 0.580      & 0.572          &         0.582  & 0.198        \\
        car                 & 0.835      & 0.830          &         0.837  & 0.412    \\
        chair               & 0.504      & 0.500          &         0.510  & 0.284    \\
        monitor             & 0.516      & 0.508          &         0.520  & 0.346    \\
        cellphone           & 0.704      & 0.689          &         0.710  & 0.497    \\
        speaker             & 0.648      & 0.659          &         0.670  & 0.505    \\
        table               & 0.536      & 0.537          &         0.540  & 0.376     \\
        \hline
    \end{tabular}
    \caption{We validate that the class conditioning is being used by the model for the GCE framework.
        Performance drops significantly when the class embedding is randomly selected.}
    \label{tab:ablate}
\end{table}

\subsubsection{Validating class prior}
It's possible that in the GCE framework as well as in \cite{wallace2019few} the trained model can learn to ignore the conditioning information. In order to validate that the class codes in the GCE framework (on which CGCE and MCCE are based) are learning meaningful priors we perform a simple ablation shown in Table~\ref{tab:ablate}. After training on base classes we randomize the class selected at test time (GCE\_rand) and observe that performance drops drastically, thus validating that the model is learning to use the class prior.  





\subsubsection{Analysis of the Compositional GCE}
\begin{figure}[h!]
\begin{center}
\centering
\begin{tabular}{cccccc }
                  GT &  \includegraphics[width=0.15\textwidth]{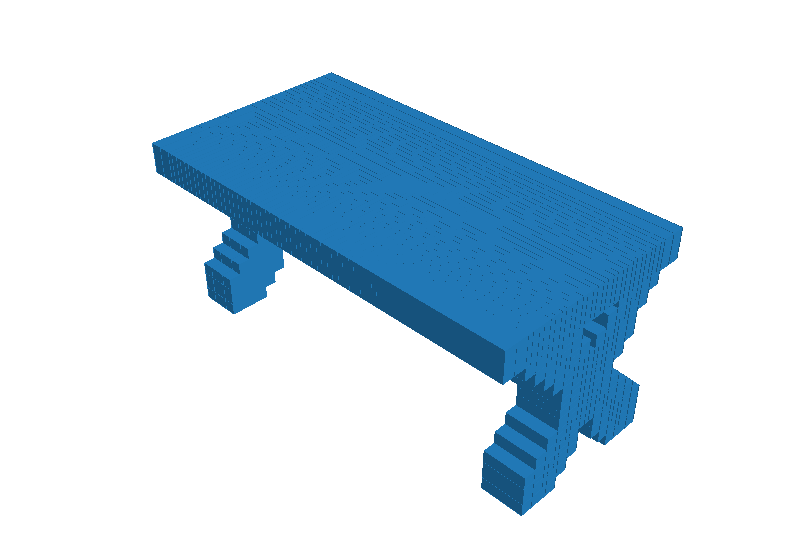} &\includegraphics[width=0.15\textwidth]{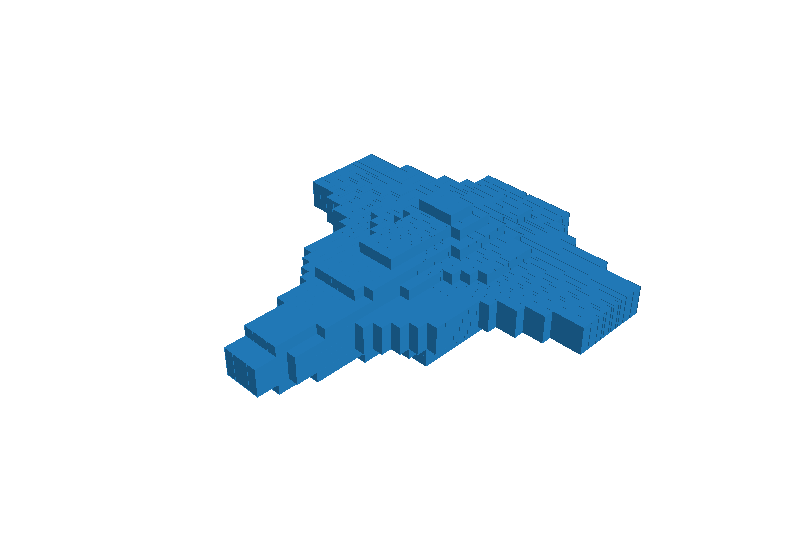} & \includegraphics[width=0.15\textwidth]{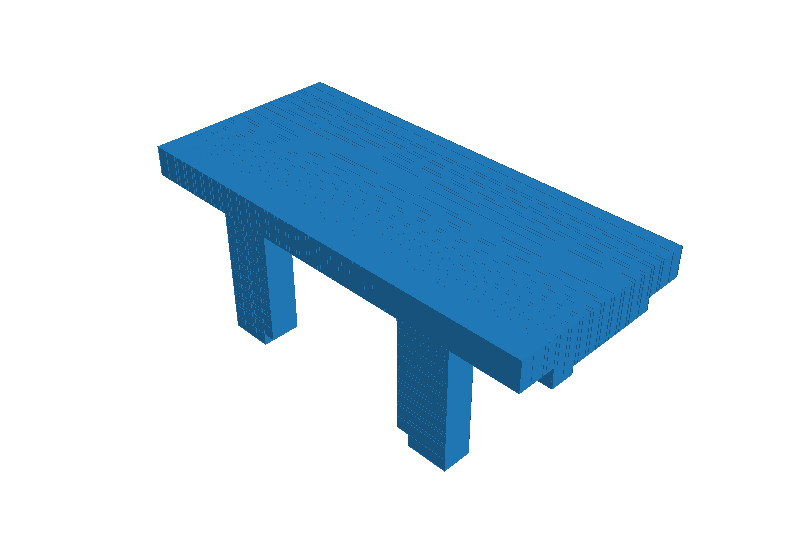} & \includegraphics[width=0.15\textwidth]{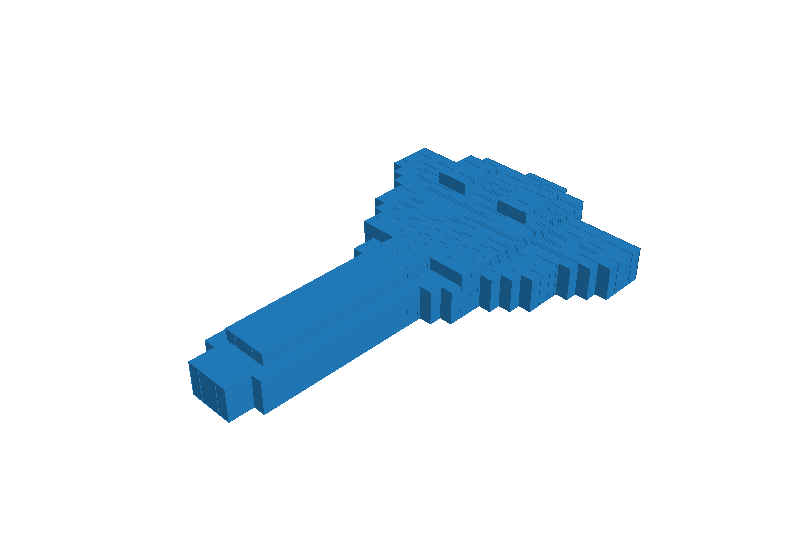} &   \includegraphics[width=0.15\textwidth]{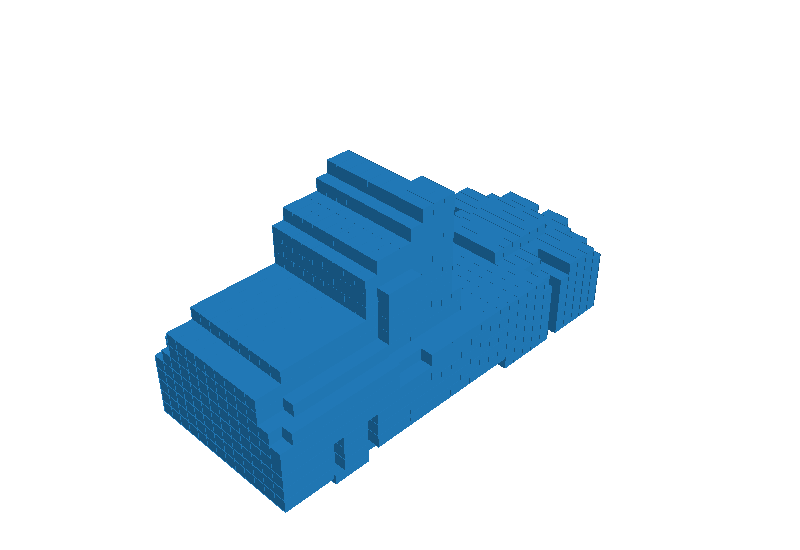}  \\
                  CGCE & \includegraphics[width=0.15\textwidth]{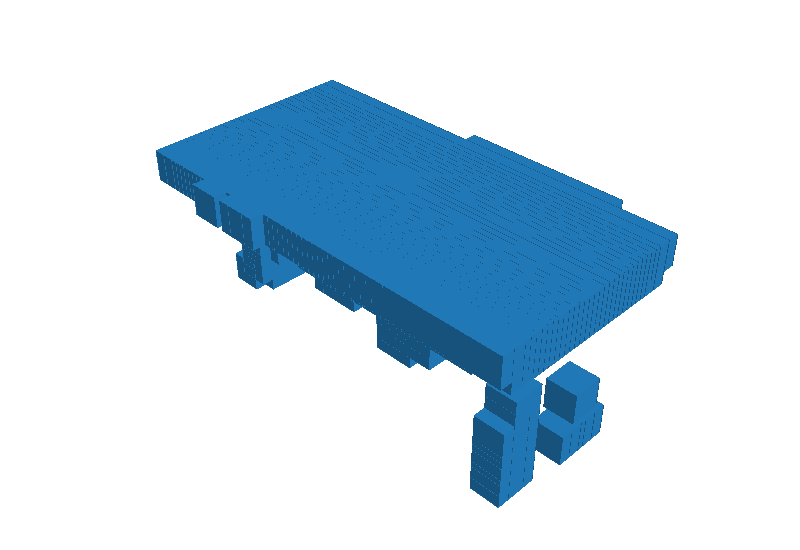} & \includegraphics[width=0.15\textwidth]{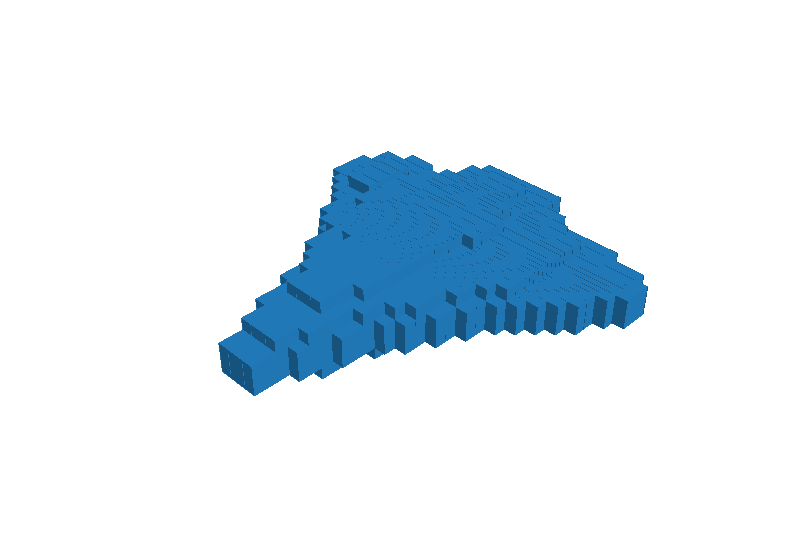} &\includegraphics[width=0.15\textwidth]{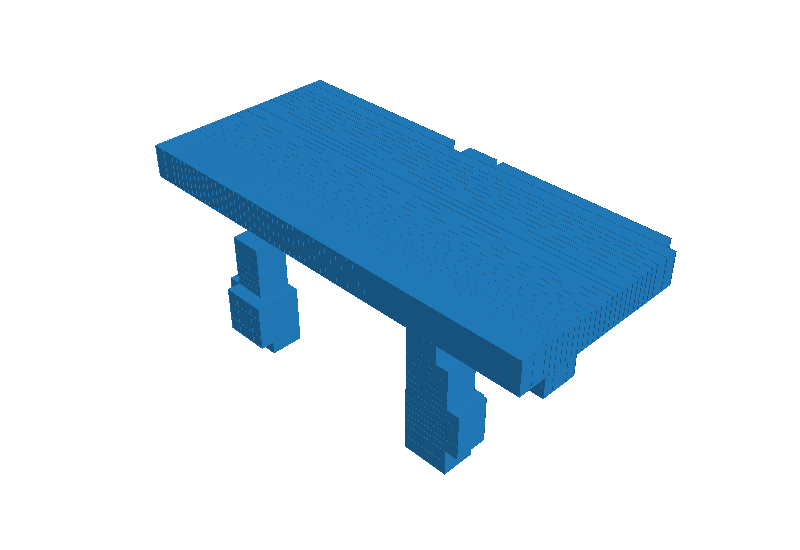} &
                            \includegraphics[width=0.15\textwidth]{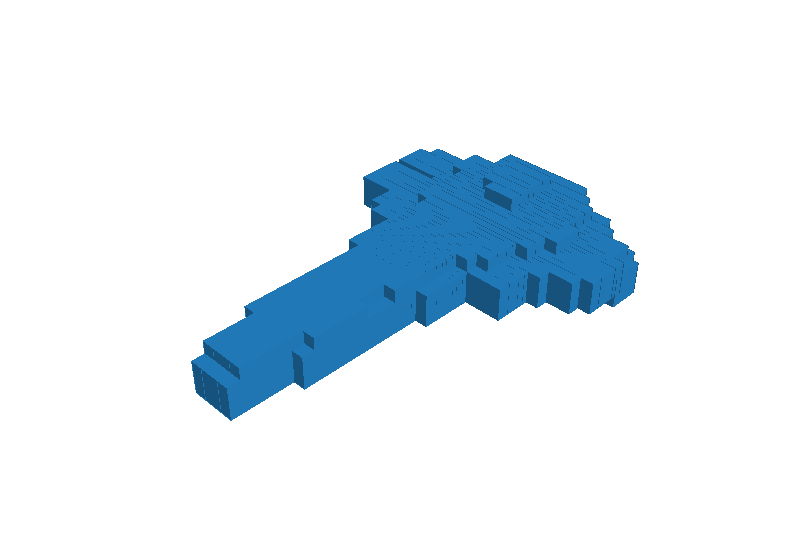} & \includegraphics[width=0.15\textwidth]{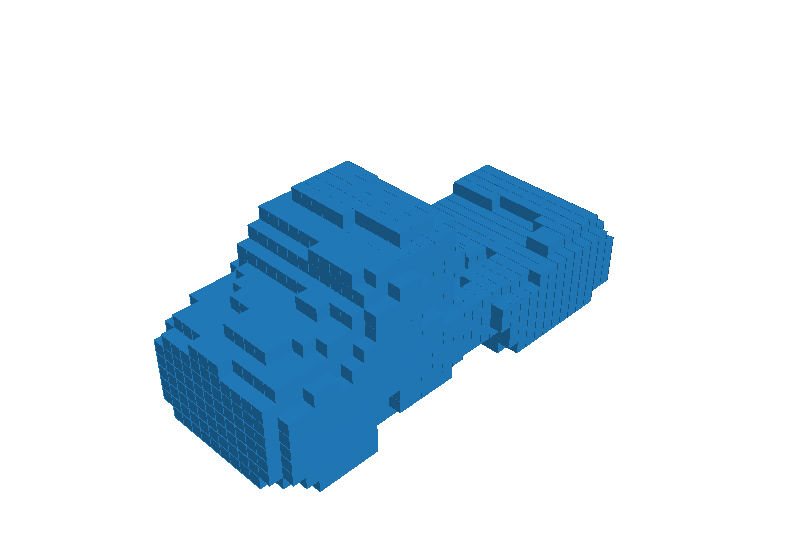} \\
                  CGCE-cb & \includegraphics[width=0.15\textwidth]{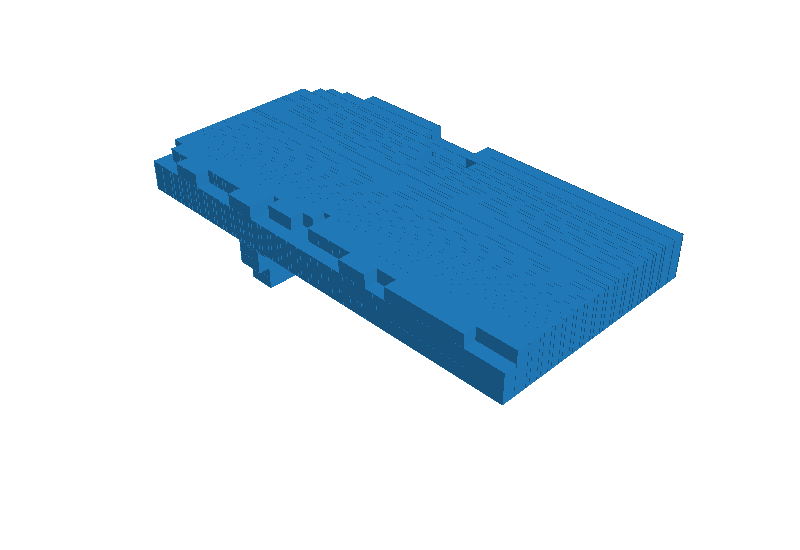} & \includegraphics[width=0.15\textwidth]{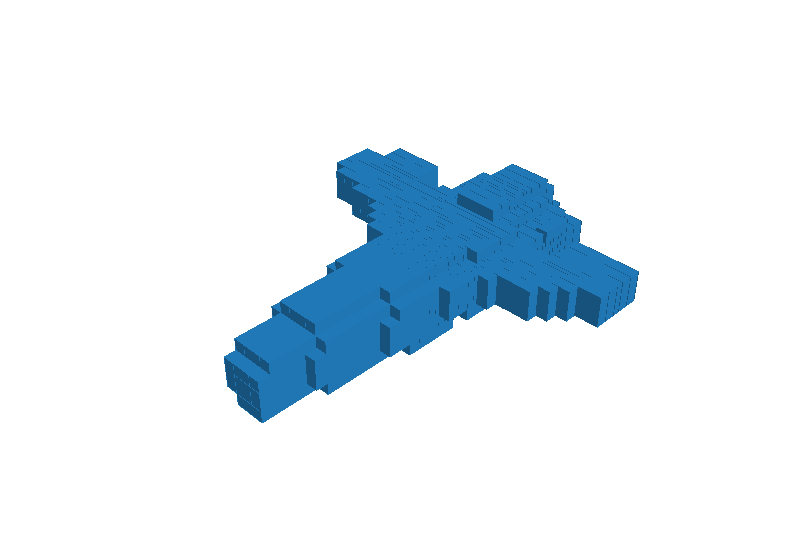}  & \includegraphics[width=0.15\textwidth]{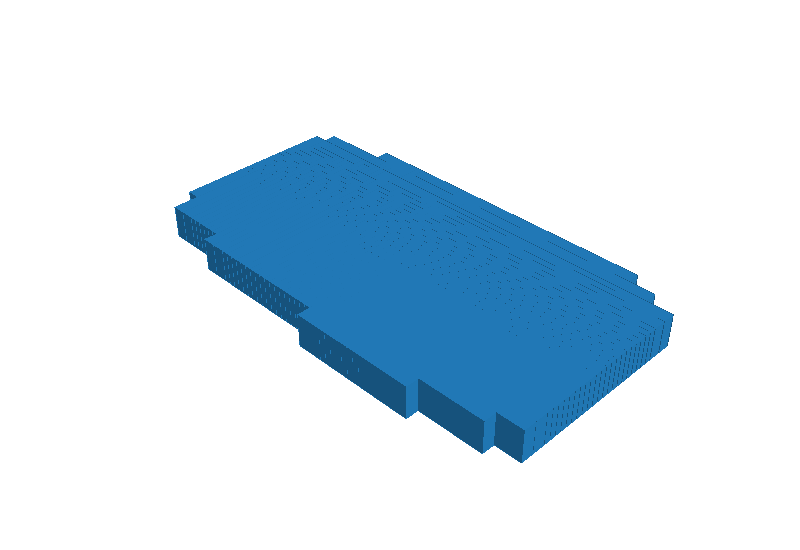} & \includegraphics[width=0.15\textwidth]{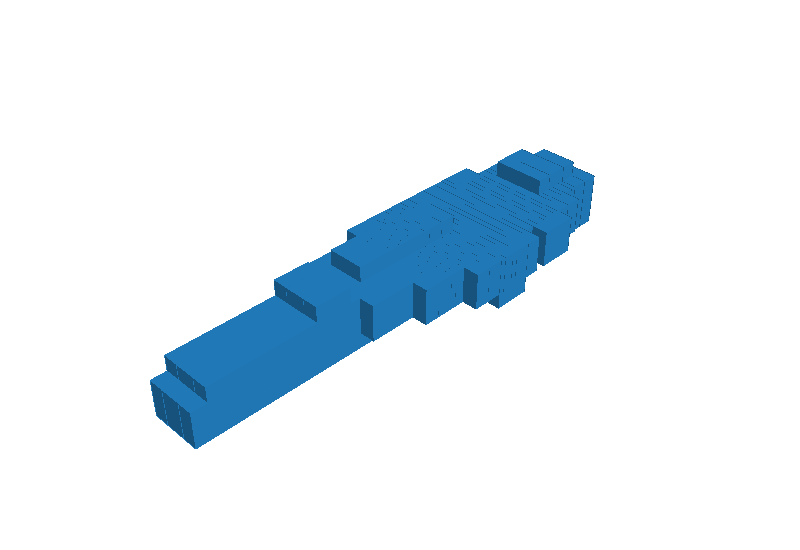} & \includegraphics[width=0.15\textwidth]{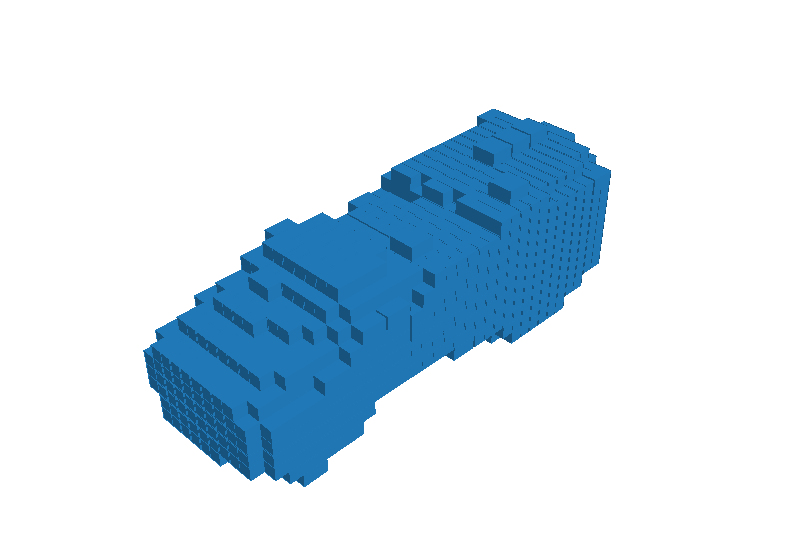}






\end{tabular}
\end{center}
\caption{ Illustration of the effect of removing one codebook from the shape conditioning vector on the shape output. The rows show ground truth shape (GT), our CGCE prediction, and the prediction ignoring one of the codebooks. We can see this results in part of the shape being removed (e.g. legs of a table or wings of an airplane).}
\label{fig:remove_code}
\end{figure}

Here we qualitatively analyze the CGCE codes learned by our model. First we attempt to visualize if codebook entries are being associated with visible concepts or parts.  We thus generate a reconstruction for an input image and then randomly remove codebook entries. The results of these experiments are illustrated in Figure~\ref{fig:remove_code}. We observe that removing some code entries can remove semantically meaningful portions of the overall reconstructed image. For example tables can be observed to lose their legs, table legs can turn into wheel like structures, and planes can be observed to lose their wings. Thus the codes to an extent exhibit some association to object parts. 


\begin{figure}
    \centering
    \includegraphics[width=0.9\textwidth]{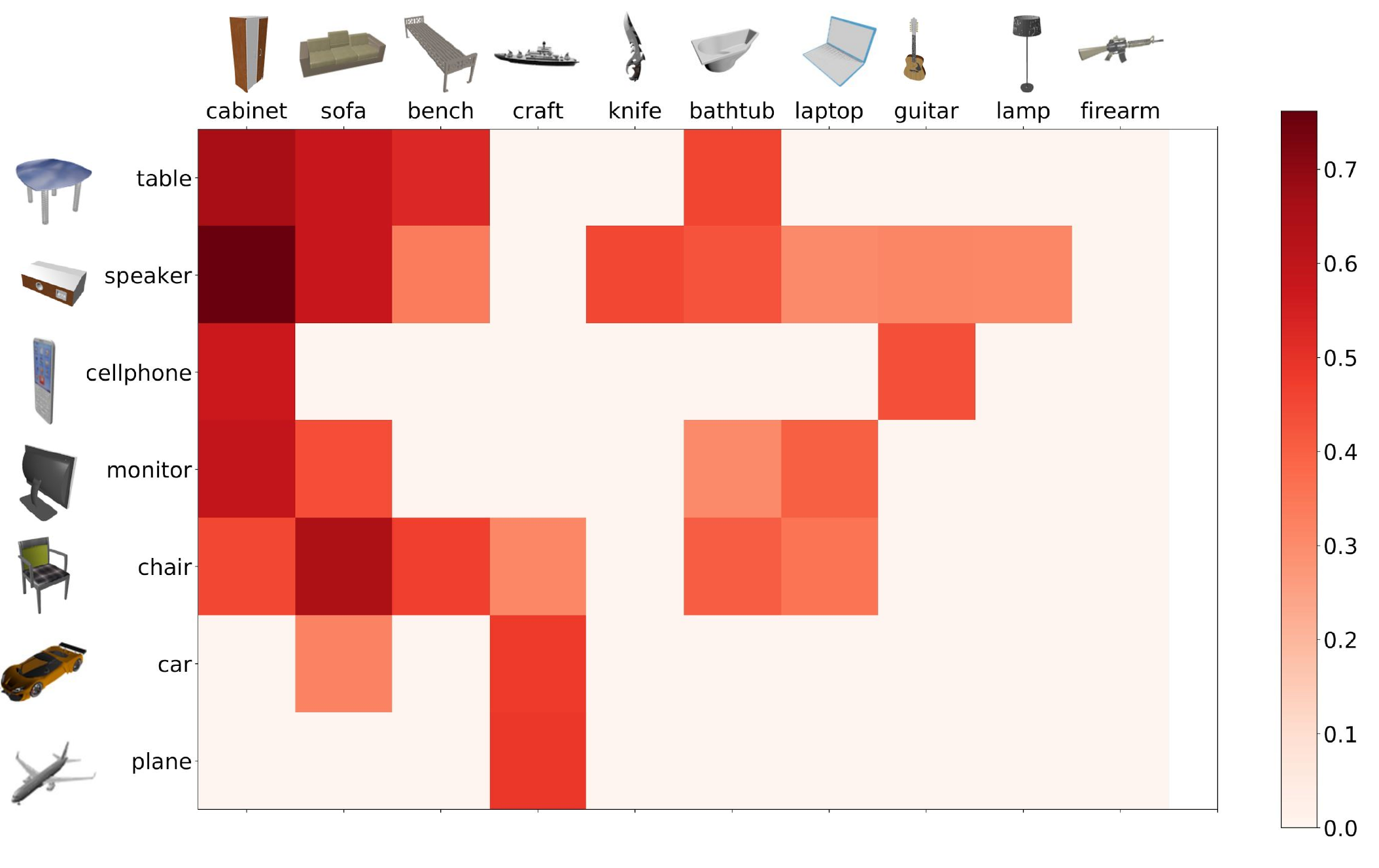}
    \caption{Proximity between base classes (y-axis) and novel classes (x-axis). Distance is measured,
        as a mean IoU of nearest neighbors.
    }
    \label{fig:proximity_matrix}
\end{figure}

We also explicitly analyze the attention learned over the codebook entries. In Figure~\ref{fig:proximity_matrix} we use the IOU based class similarity metric to associate each novel class to its closest base classes. 
We observe that when a novel classes attends to the same code as a base class, this will tend to align with our proximity metric, but not for all code-books. Indeed one would expect that classes sharing similar structures will have many similar concepts, but differ in some. The full visualization of these experiments is included in the supplementary materials.  
\subsubsection{Qualitative results}
Finally we visualize our reconstructions as compared to those of \cite{wallace2019few} in the 25-shot case. We observe that the numerical performance gains can also be seen qualitatively in the reconstructions.  

\begin{figure}[h!]
\centering
\begin{tabular}{cccccc|c|c}
\hline
                 2D view& Zero-Shot &      Wallace       & CGCE            &  GT              \\
        \hline
          \includegraphics[width=0.13\textwidth]{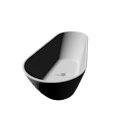} &  \includegraphics[width=0.15\textwidth]{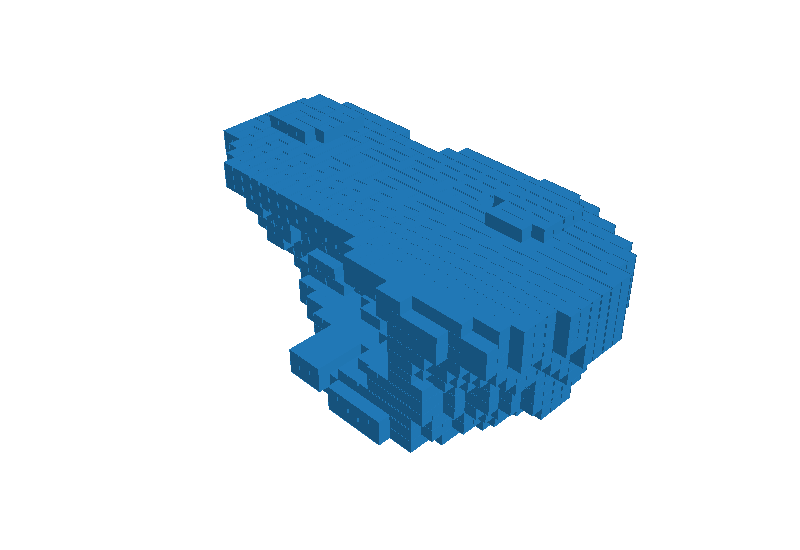}           &    \includegraphics[width=0.15\textwidth]{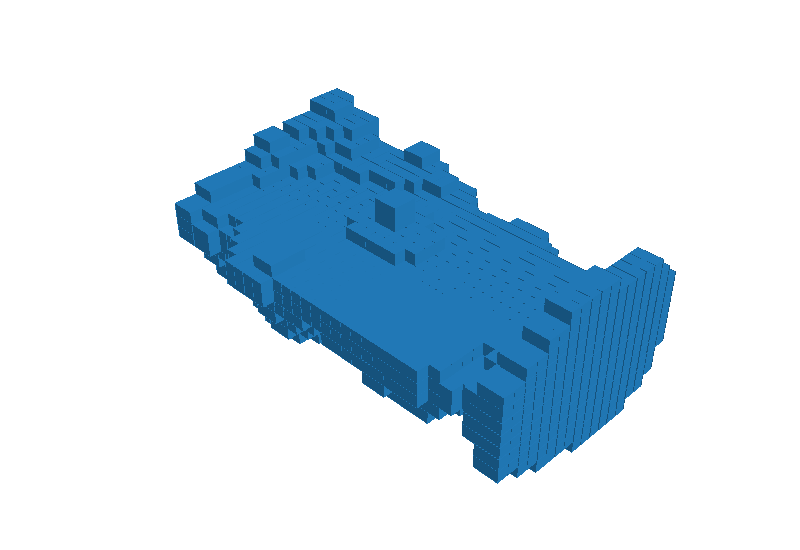}             &   \includegraphics[width=0.15\textwidth]{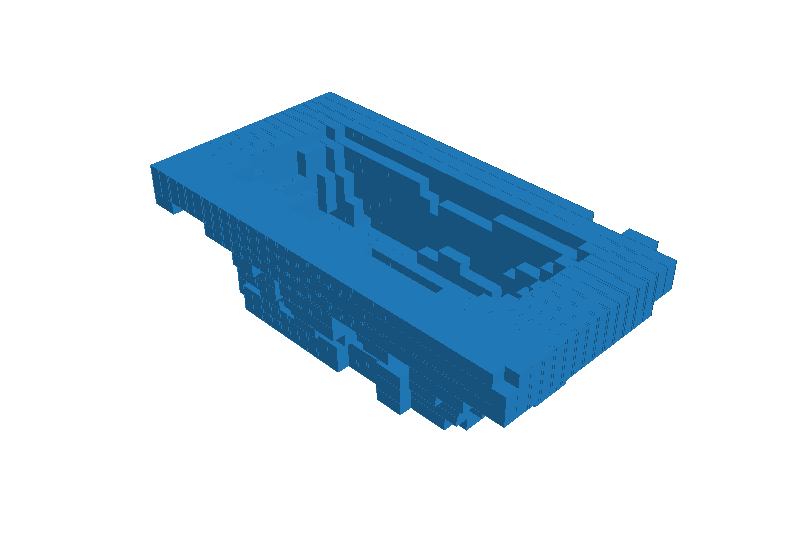}              &     \includegraphics[width=0.15\textwidth]{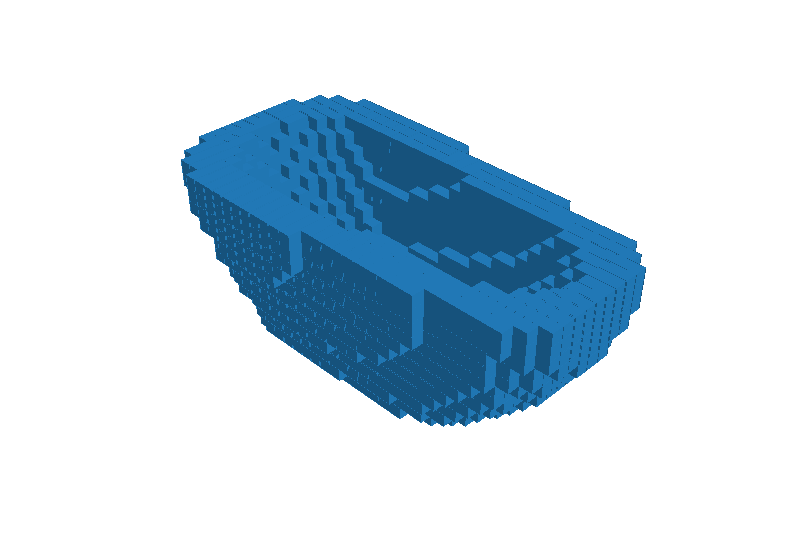}               \\
           \includegraphics[width=0.13\textwidth]{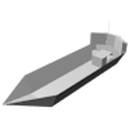} &  \includegraphics[width=0.15\textwidth]{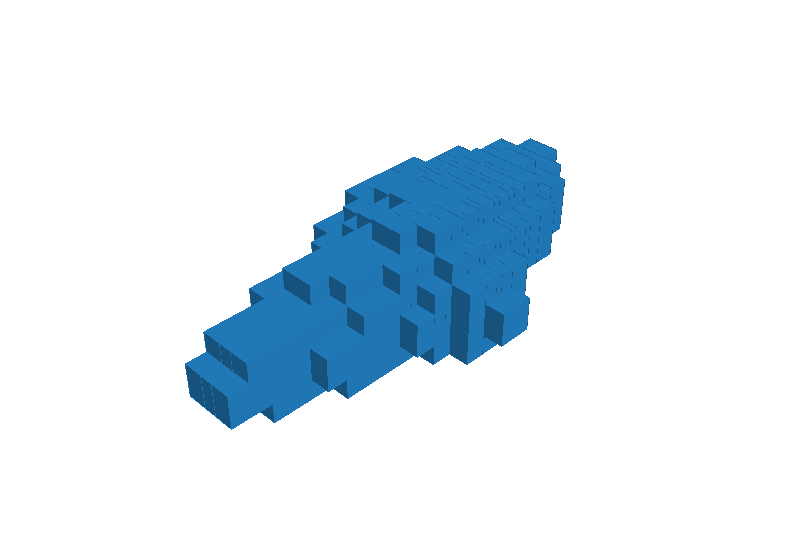}           &    \includegraphics[width=0.15\textwidth]{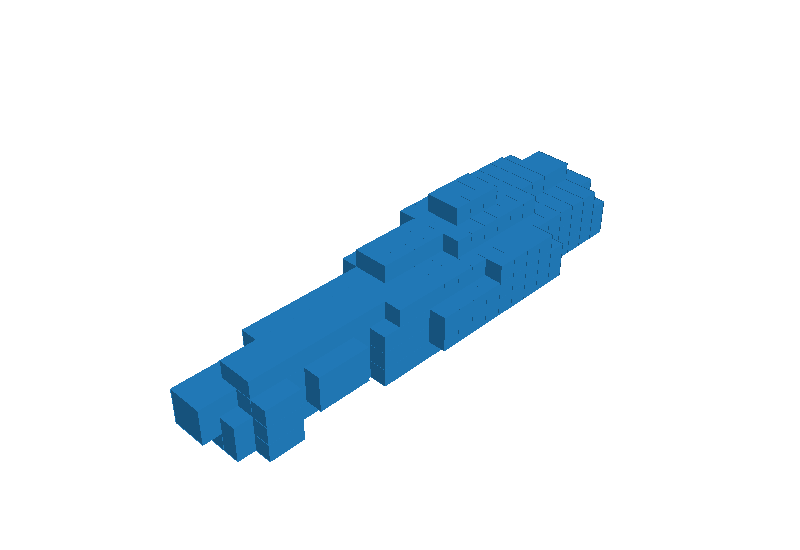}             &   \includegraphics[width=0.15\textwidth]{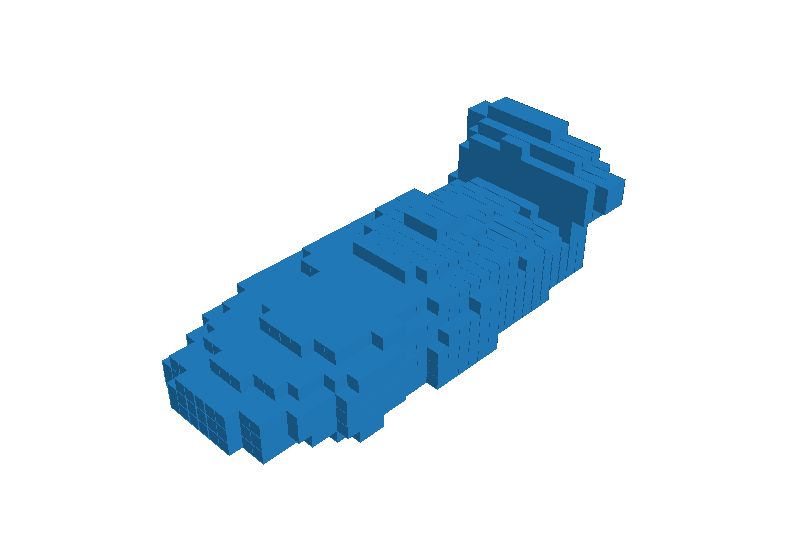}              &     \includegraphics[width=0.15\textwidth]{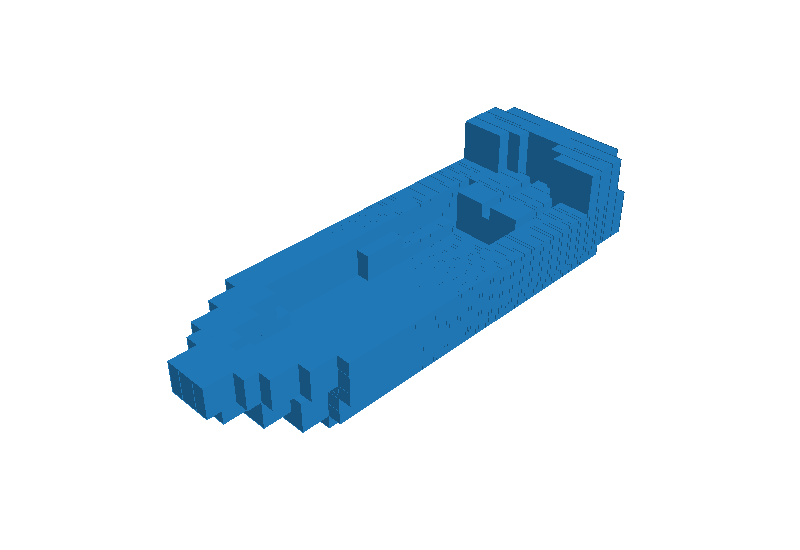}               \\ 
          \includegraphics[width=0.13\textwidth]{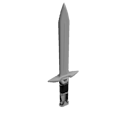} &  \includegraphics[width=0.15\textwidth]{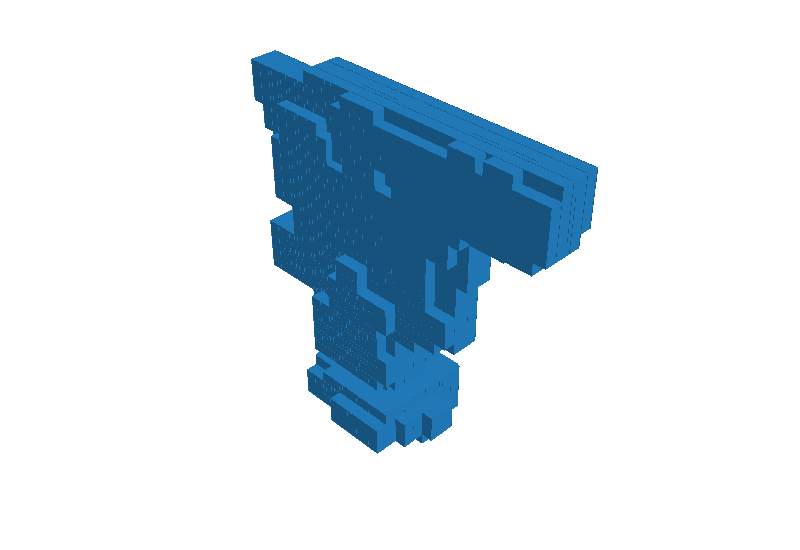}           &    \includegraphics[width=0.15\textwidth]{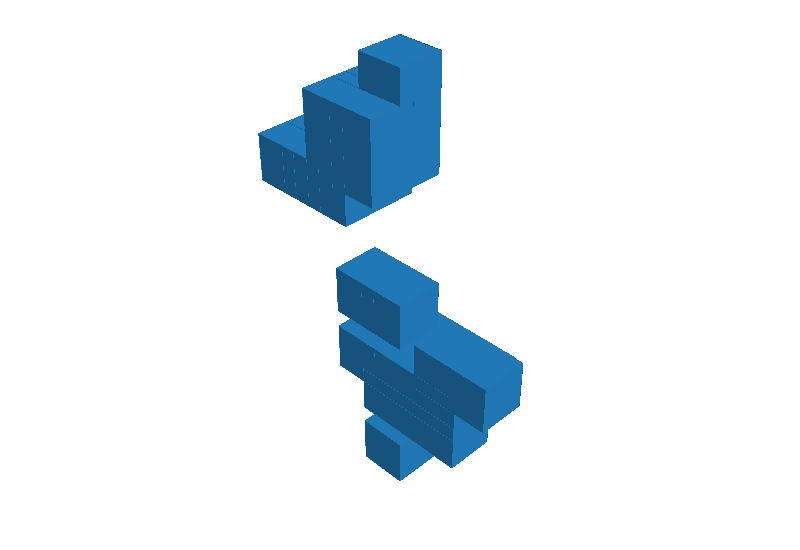}             &   \includegraphics[width=0.15\textwidth]{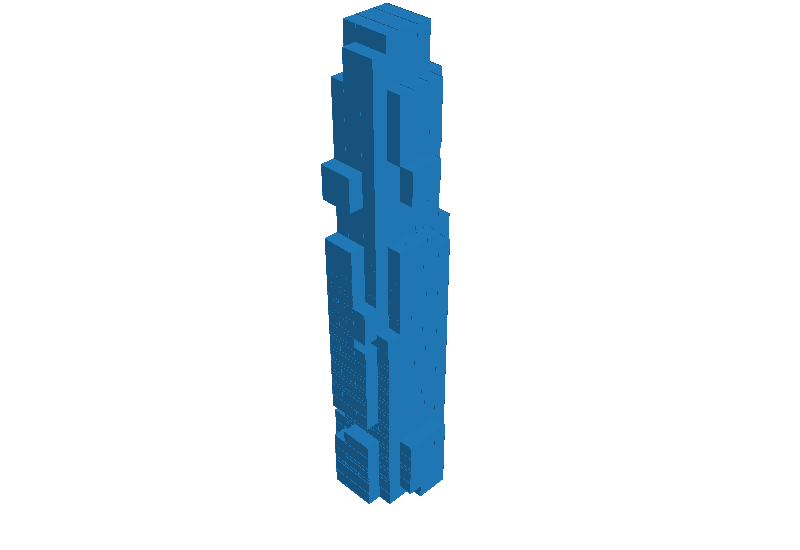}              &     \includegraphics[width=0.15\textwidth]{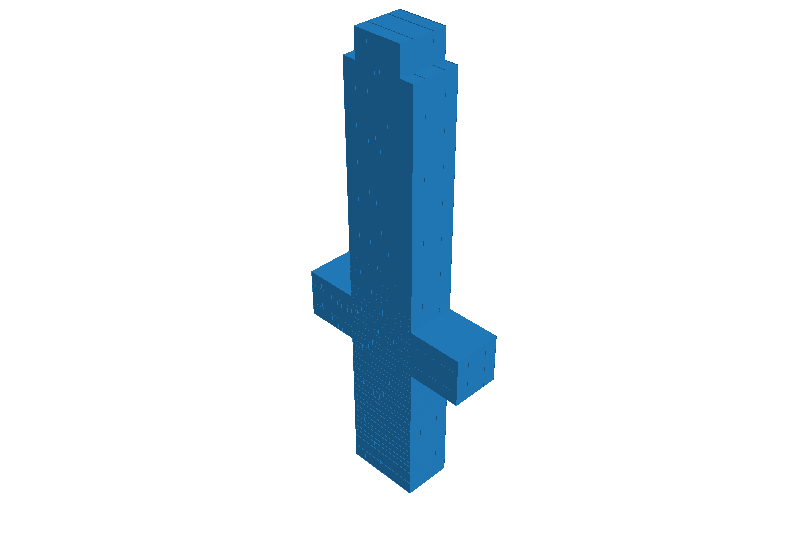}               \\

        \hline
\end{tabular}
\caption{Qualitative analysis on 3 different examples using novel classes with 25-shots. Predictions from different models are shown as well as the ground truth (GT). We observe that our model exhibits qualitatively better reconstructions than Wallace or the Zero-Shot baseline. 
}
\label{fig:1}
\end{figure}

\section{Conclusions}
We have highlighted that few shot generalization can be an excellent benchmark for studying 3-D deep learning models and their ability to generalize about object shapes. We addressed several key weaknesses of models in this setting, particularly showing we can deal with intra-class variability and that inducing compositional and multi-scale priors can improve greatly the performance on this benchmark. Future work in this area can aim to study how various shape representations which have been developed for 3-D object reconstruction perform on this benchmark, and whether their inductive biases can further improve generalization, particularly for higher resolution shapes.  
\clearpage
%
%
\bibliographystyle{splncs04}
\bibliography{egbib}
\appendix
\section{Appendix}

We provide additional material to supplement our work. Appendix \ref{reimp} verifies the accuracy of our re-implementation of \cite{wallace2019few}, which is, to our knowledge, the only pre-existing work on few-shot 3D reconstruction. 
In Appendix \ref{baseclass}, we report performance on base classes for our three considered methods and Wallace et al. \cite{wallace2019few}. Appendix \ref{amaps} shows learned attention maps obtained using the CGCE model and analyses similarities across classes. Finally, we provide qualitative examples in Appendix \ref{qual}.

\subsection{Verifying Implementation of Wallace et al.~\cite{wallace2019few}}
\label{reimp}

In this section we validate that our re-implementation of \cite{wallace2019few} is correct. In Table~\ref{table:wallace} we observe performance to be very similar to the numbers reported in \cite{wallace2019few}, with small variations that can be reasonably attributed to random initializations. Base class performance is not reported per class in \cite{wallace2019few}. Note that in the main paper we report the results obtained using our implementation (Wallace(ours)), including the results on classes not attempted in \cite{wallace2019few}.
\begin{table}[h!]
    \centering
    \begin{tabular}{|c|c|c|}
        \hline
        cat              & Wallace  \cite{wallace2019few}   & Wallace(ours)                                             \\
        \hline
        \multicolumn{3}{|c|}{ \textbf{base}}  \\
        \hline
        plane            & N/A             & 0.57                      \\
        car              & N/A             & 0.84                      \\
        chair            & N/A             & 0.49                      \\
        monitor          & N/A             & 0.50                      \\
        cellphone        & N/A             & 0.74                      \\
        speaker          & N/A             & 0.66                        \\
        table            & N/A             & 0.52                        \\
        \hline
        mean\_base       & 0.62           & 0.62                       \\
        \hline
        \multicolumn{3}{|c|}{ \textbf{novel}}  \\
        \hline

        bench            & 0.37 (0\%)     & 0.37 (0\%)                            \\
        cabinet          & 0.66 (0\%)     & 0.69 (0\%)                            \\
        lamp             & 0.19 (5\%)     & 0.20 (5\%)                            \\
        firearm          & 0.19 (58\%)    & 0.21 (58\%)                                    \\
        couch            & 0.52 (4\%)     & 0.54 (4\%)                           \\
        watercraft       & 0.38 (15\%)    & 0.33 (16\%)                                      \\
        \hline
        \hline
        mean\_novel      & 0.39           & 0.39                                     \\
        \hline
    \end{tabular}
    \caption{Comparison of Wallace et al. \cite{wallace2019few} and our re-implementation validates our experiments. 
    N/A means numbers were not reported in \cite{wallace2019few}. 
    Numbers in brackets indicate percentage improvement over baseline.}
    \label{table:wallace}
\end{table}

\subsection{Base Class Performance} \label{baseclass}
We report in Table~\ref{table:base} the performance on base classes for all methods. Note that performance is similar amongst methods. This is consistent with our observation that nearest neighbor can solve the problem when large number of classes is provided. Thus most learning methods with enough capacity can expect to obtain similar performance. On the other hand, as shown in the main paper, novel class performance is improved for our proposals (GCE,CGCE,MCCE) demonstrating they generalize better about shapes.  

\begin{table}[h!]
    \centering
    \begin{tabular}{|c|c|c|c|c|}
        \hline
        cat         &                             Wallace & GCE    & CGCE  & MCCE        \\
        \hline
        \multicolumn{5}{|c|}{ \textbf{base}}  \\
        \hline
        plane                       &             0.57    & 0.58   & 0.59  & 0.59        \\
        car                         &             0.84    & 0.84   & 0.84  & 0.84        \\
        chair                       &             0.49    & 0.51   & 0.49  & 0.50         \\
        monitor                     &             0.50    & 0.52   & 0.51  & 0.52        \\
        cellphone                   &             0.74    & 0.71   & 0.69  & 0.71        \\
        speaker                     &             0.66    & 0.67   & 0.66  & 0.66         \\
        table                       &             0.52    & 0.54   & 0.54  & 0.53         \\
        \hline
        mean\_base                  &             0.62    & 0.62   & 0.62  & 0.62       \\
        \hline
    \end{tabular}
    \caption{Results on base classes for all methods. All methods perform similarly which is consistent with our    
        observation that for large number of classes the problem is reduced to a simple nearest neighbor search.
    }
\label{table:base}
\end{table}







\section{Compostional GCE Further Analysis: Attention Maps}
\label{amaps}

As described in the main text, a learned attention vector $\boldsymbol{\alpha}_i$ selects the most relevant codes from each of the 5 available codebooks. We visualize these selections for each category as a heat map in Figure~\ref{fig:heat_map}. Note that we have previously obtained a similarity metric between classes (using nearest neighbor proximity) as shown in Figure 7 in the main text. In Table~\ref{table:similarity} we further illustrate some pairs which show high similarity (as per our proximity metric). We observe in Figure~\ref{fig:heat_map} that \textit{similar classes will often share codes}.  We further illustrate this by selecting 3 pairs of similar categories and 3 pairs of distant categories (see Table~\ref{table:similarity} and Figure~\ref{fig:total_map}). 
Indeed this shows that CGCE model is learning to assign general structure to each class which can be reused in similar classes. As expected, however, not all codebooks for similar classes share exactly the same codes, thus they can learn distinctions across classes.

\begin{table}[h!]
    \centering
    \def\w{0.2}
    \begin{tabular}{|c||c|c|          }
        \hline
        object            & similar object    &     distant object  \\
        \hline
        \includegraphics[width=\w\textwidth]{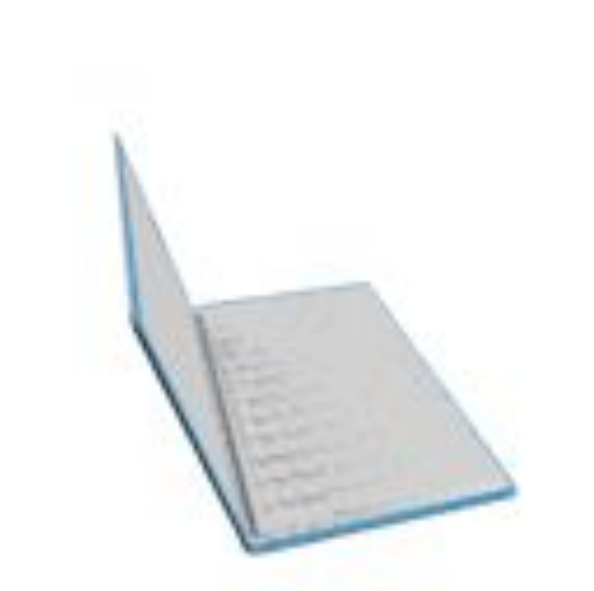}   &
        \includegraphics[width=\w\textwidth]{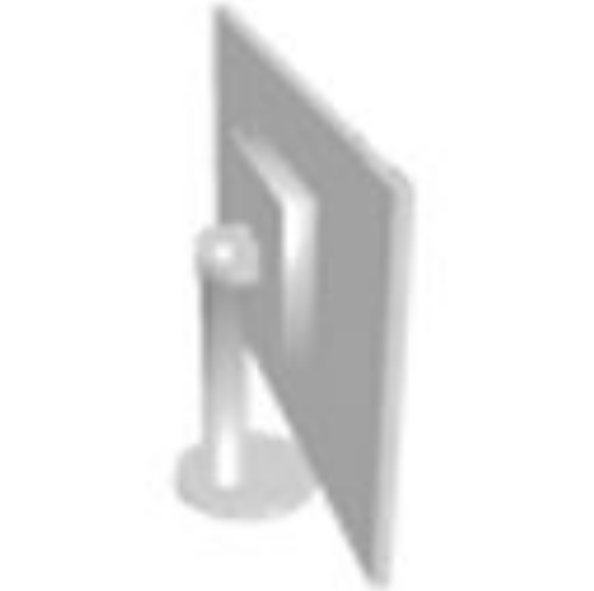}  &
        \includegraphics[width=\w\textwidth]{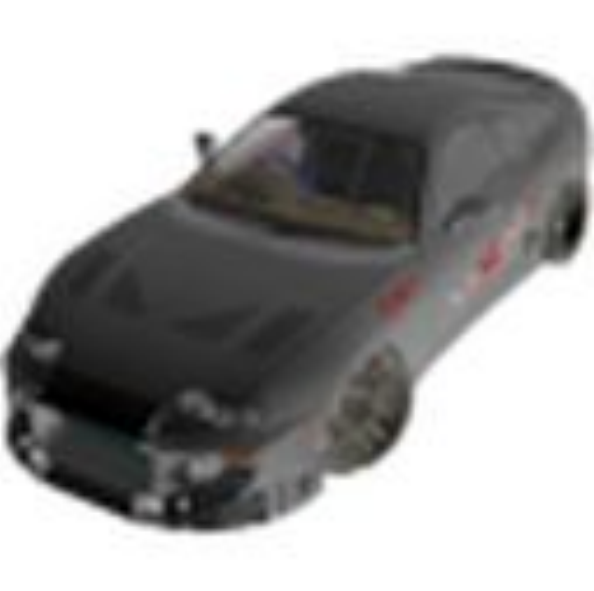} \\ 
        \hline
        \includegraphics[width=\w\textwidth]{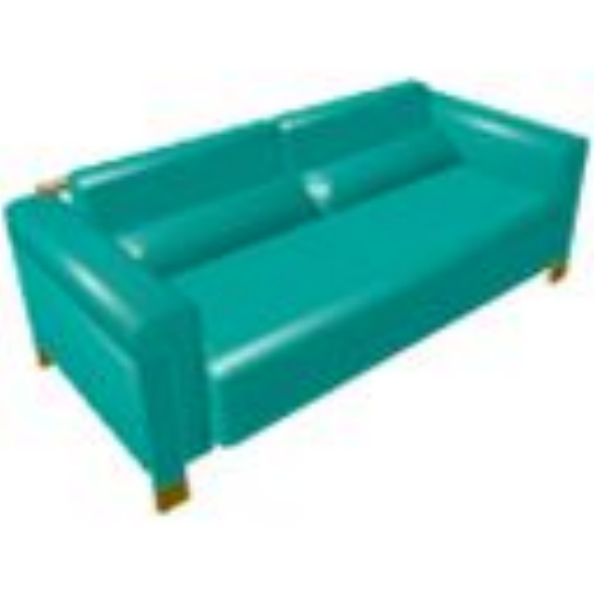}   &
        \includegraphics[width=\w\textwidth]{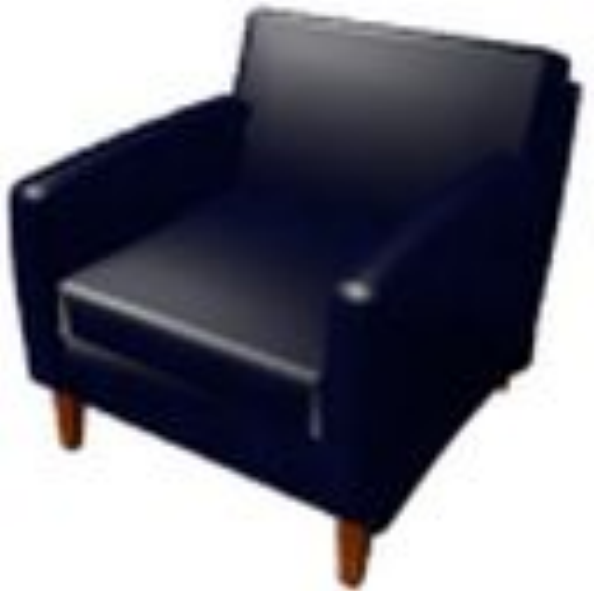}  &
        \includegraphics[width=\w\textwidth]{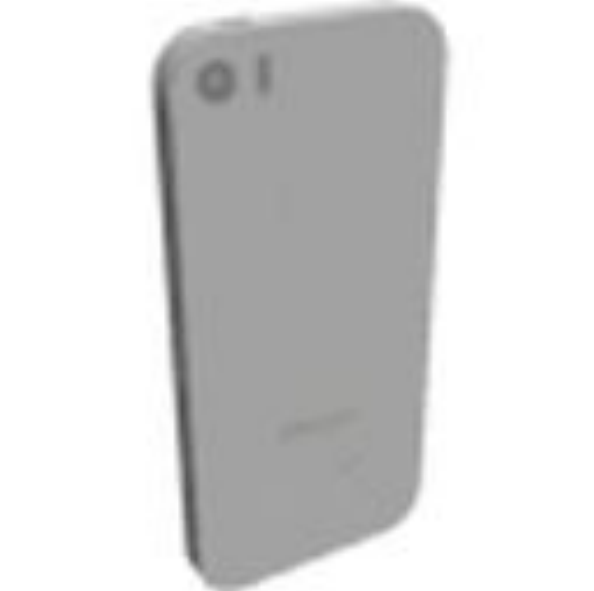} \\ 
        \hline
        \includegraphics[width=\w\textwidth]{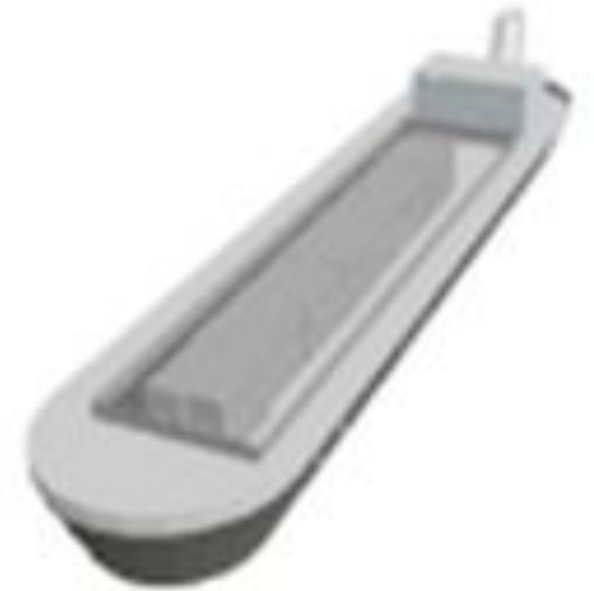}   &
        \includegraphics[width=\w\textwidth]{figures/map/car.pdf}  &
        \includegraphics[width=\w\textwidth]{figures/map/monitor.pdf} \\ 
        \hline
    \end{tabular}
    \caption{ We illustrate the categories found to be similar and dissimilar 
        (based on our nearest neighbor proximity metric). 
        For example, the first row shows that laptops are similar to monitors but distant from cars. 
        Second row indicates sofas are similar to chairs but distant from phones. 
        Third row shows watercrafts are similar to cars but distant from monitors. 
    }
    \label{table:similarity}
\end{table}

\begin{figure}
    \centering
    \def\w{0.7}
    \includegraphics[width=\w\textwidth]{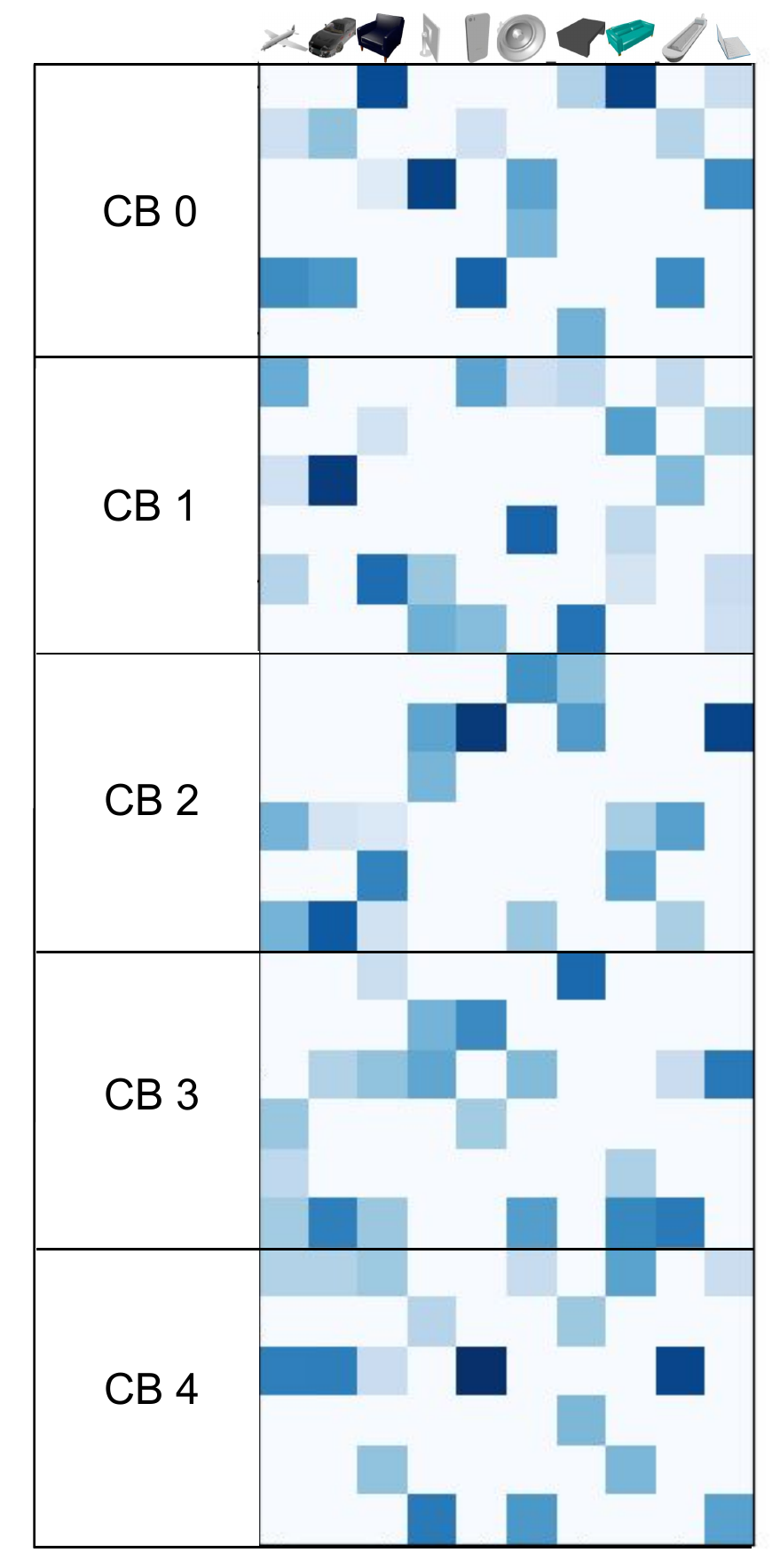}
    \caption{Attention heat map depicting code selection for all base classes (columns 1-7) and 3 novel ones (columns 8-10). We have used 5 codebooks each one having 6 codes. Darker squares indicate higher attention given to a particular code. Note each codebook J is indicated by CB J.}
    \label{fig:heat_map}
\end{figure}

 \begin{figure}
\centering
\includegraphics[width=0.98\textwidth]{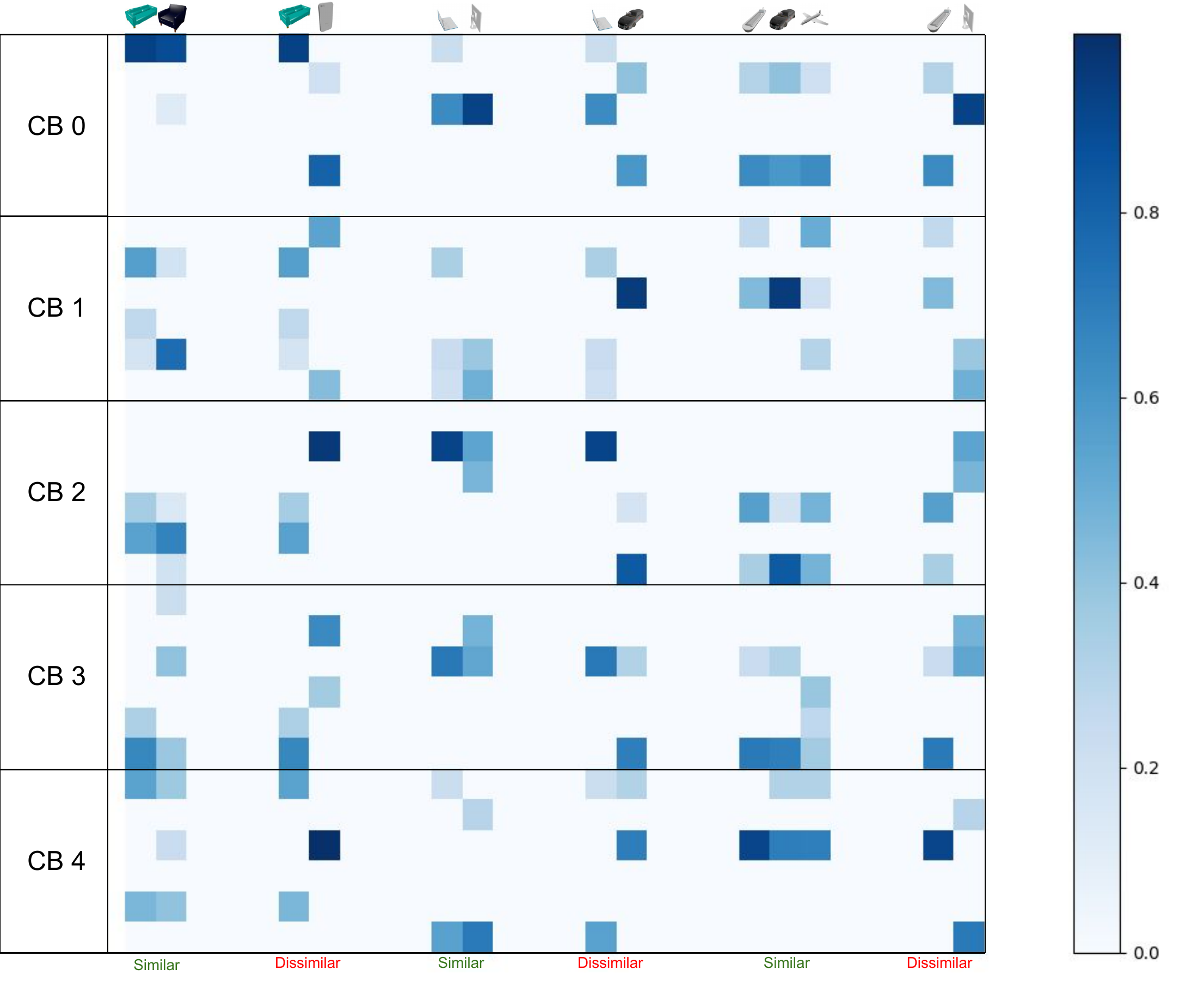}
\caption{Attention heat map of similar and distant categories. Columns feature 3 similar cases:
(sofa and chair), (laptop and monitor), (watercraft and plane and car),
and 3 distant ones:
(sofa and phone), (laptop and car), and (watercraft and monitor). 
One can see that for similar categories, similar codes are chosen as opposed to distant ones.}
\label{fig:total_map}
\end{figure}

\section{Additional Qualitative Examples}
\label{qual}

We provide more visualizations of our reconstructions as compared to Zero-Shot baseline and Wallace which demonstrate higher quality predictions obtained using our proposed method.

\begin{figure}[h!]
\centering
\begin{tabular}{ccccc}
\hline
2D view& Zero-Shot &      Wallace       & CGCE            &  GT              \\
\hline
\includegraphics[width=0.14\textwidth]{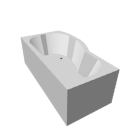} &
\includegraphics[width=0.20\textwidth]{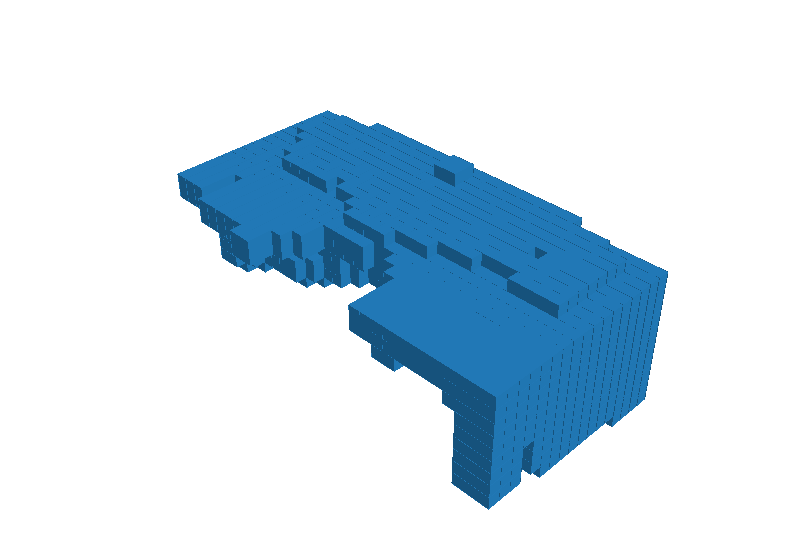} &
\includegraphics[width=0.20\textwidth]{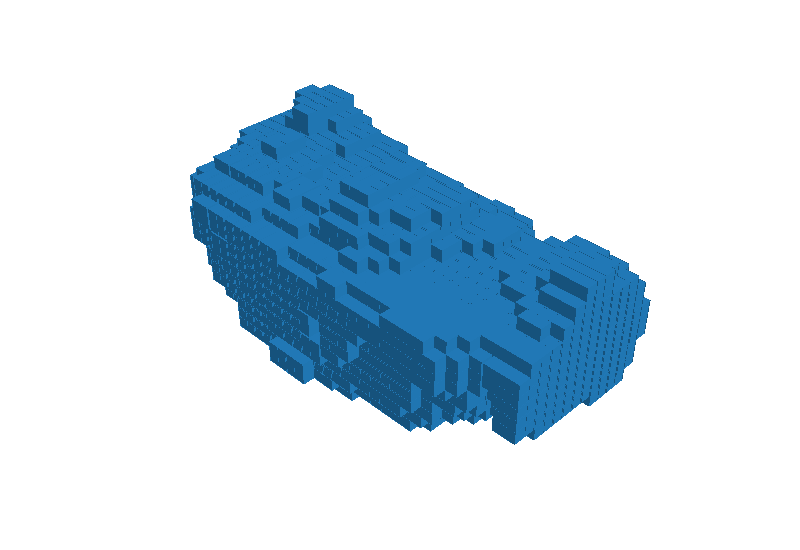} &
\includegraphics[width=0.20\textwidth]{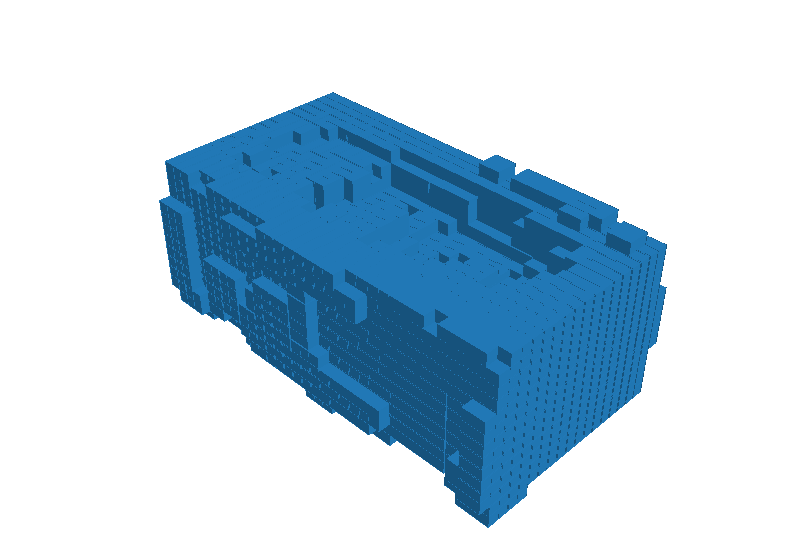} &
\includegraphics[width=0.20\textwidth]{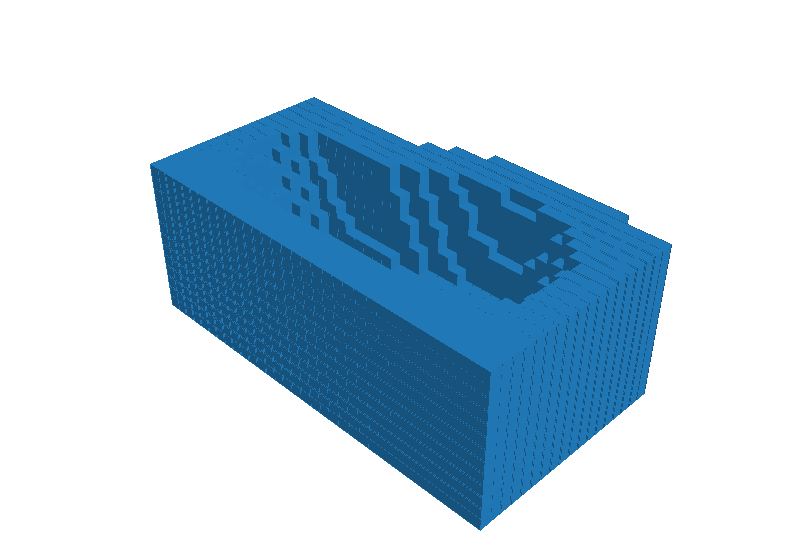} \\



\includegraphics[width=0.14\textwidth]{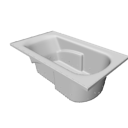} &
\includegraphics[width=0.20\textwidth]{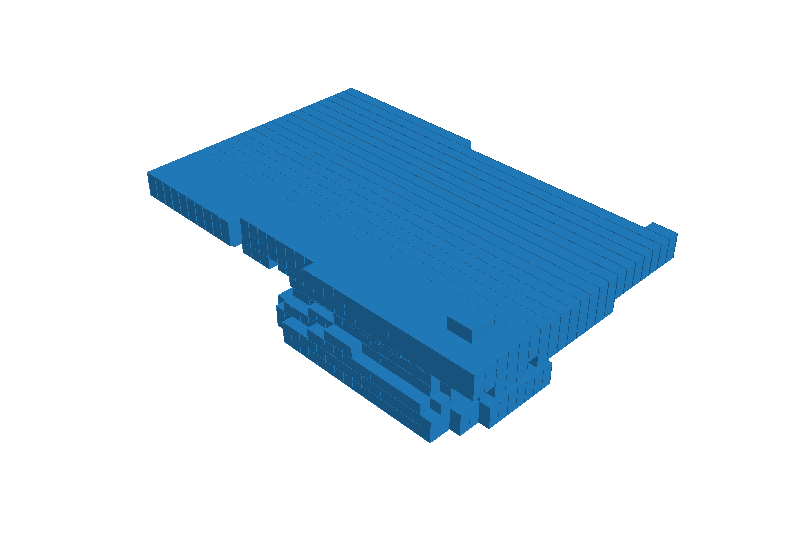} &
\includegraphics[width=0.20\textwidth]{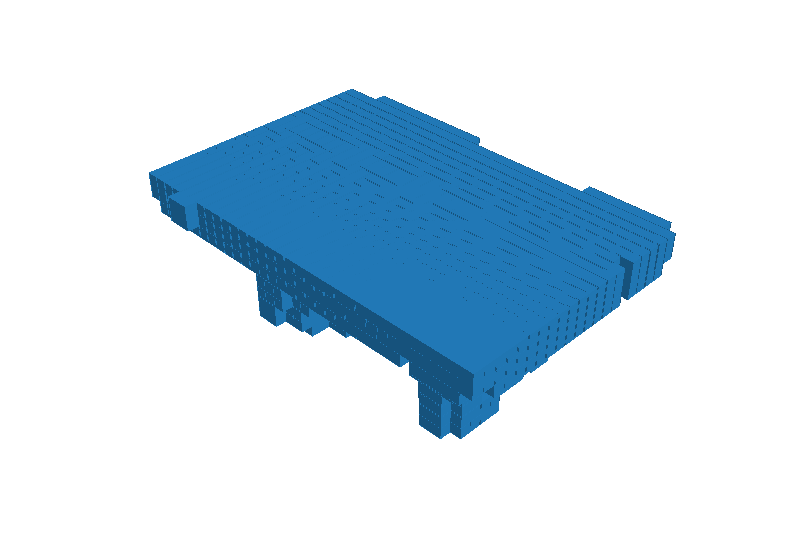} &
\includegraphics[width=0.20\textwidth]{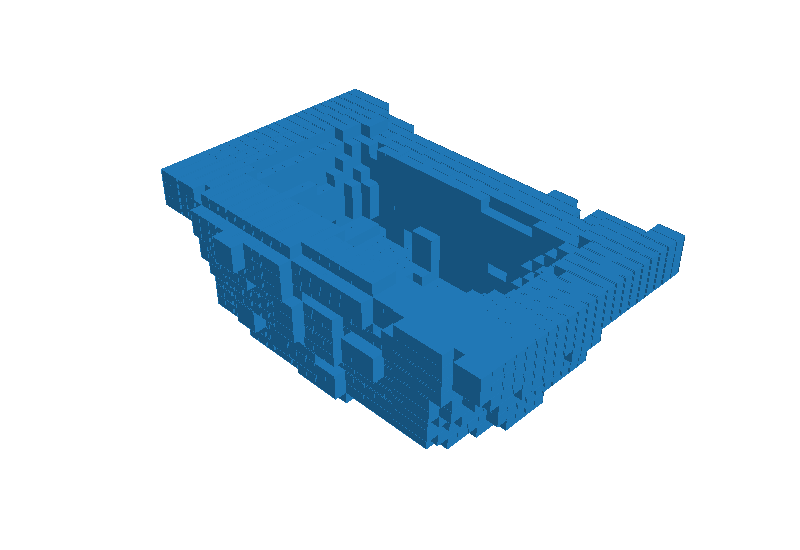} &
\includegraphics[width=0.20\textwidth]{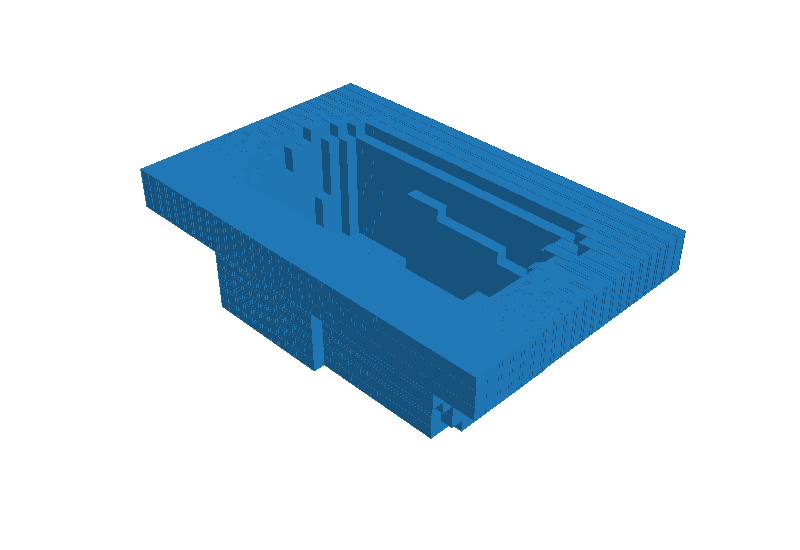} \\

\includegraphics[width=0.14\textwidth]{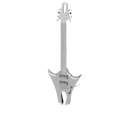} &
\includegraphics[width=0.20\textwidth]{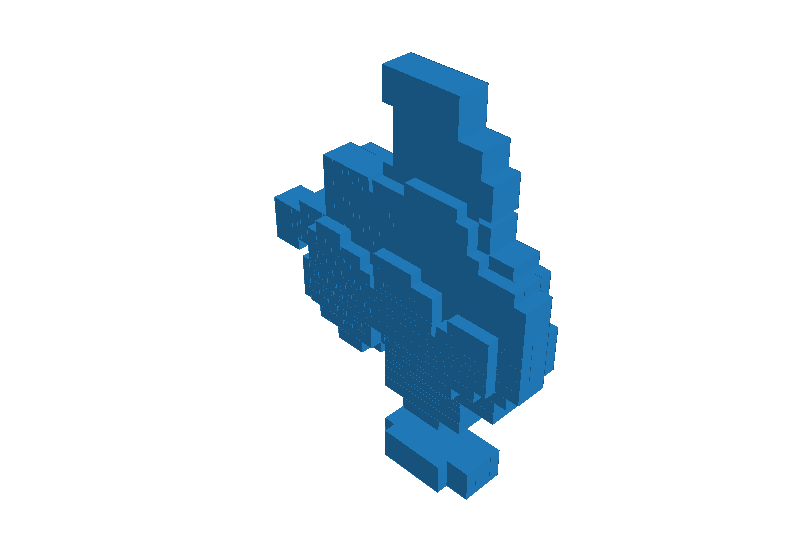} &
\includegraphics[width=0.20\textwidth]{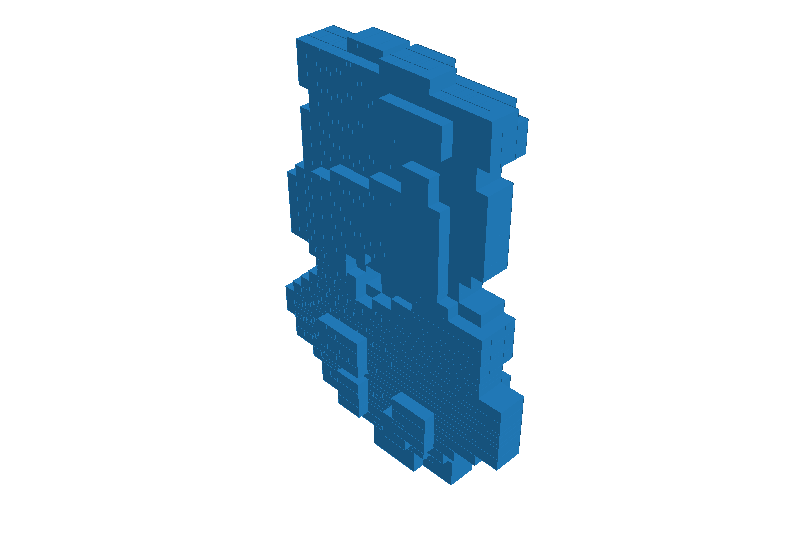} &
\includegraphics[width=0.20\textwidth]{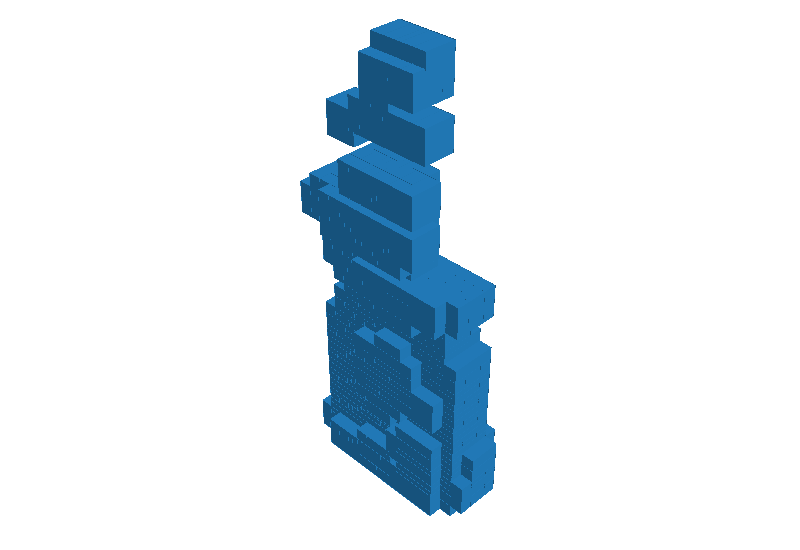} &
\includegraphics[width=0.20\textwidth]{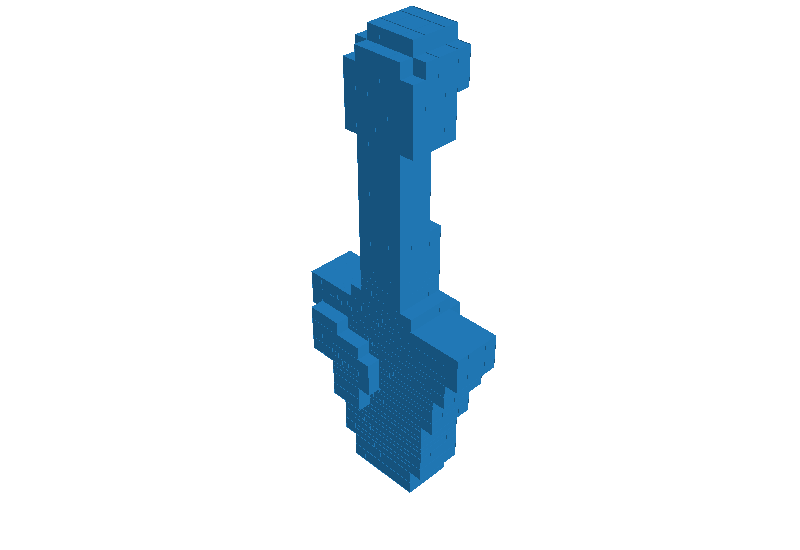} \\



\includegraphics[width=0.14\textwidth]{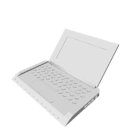} &
\includegraphics[width=0.20\textwidth]{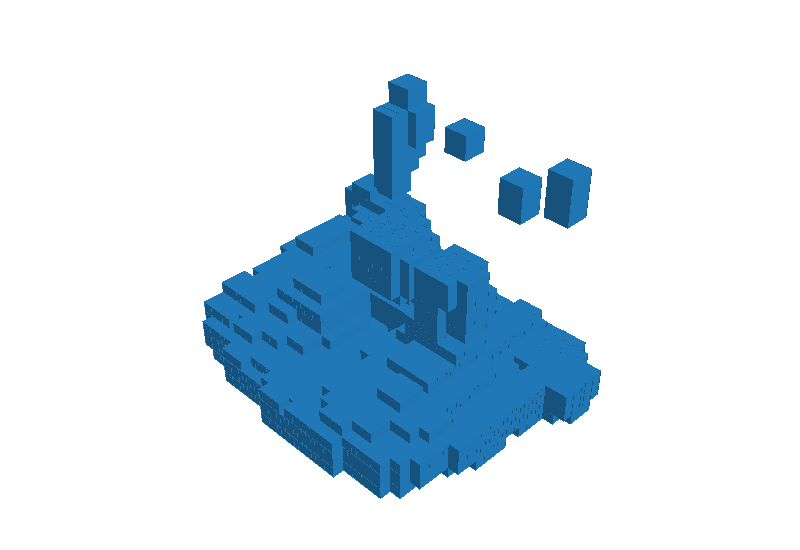} &
\includegraphics[width=0.20\textwidth]{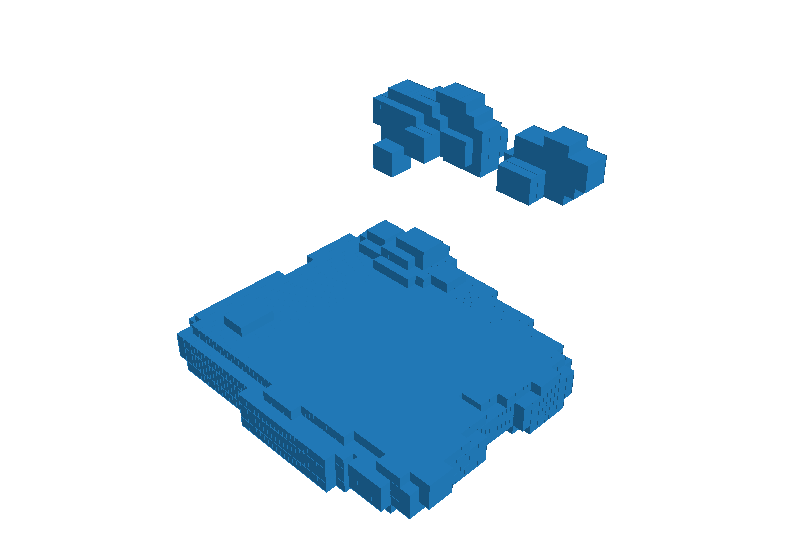} &
\includegraphics[width=0.20\textwidth]{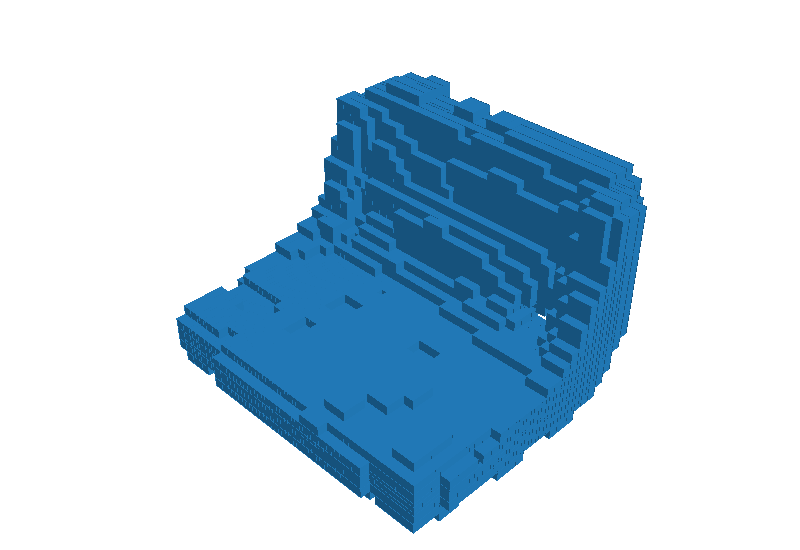} &
\includegraphics[width=0.20\textwidth]{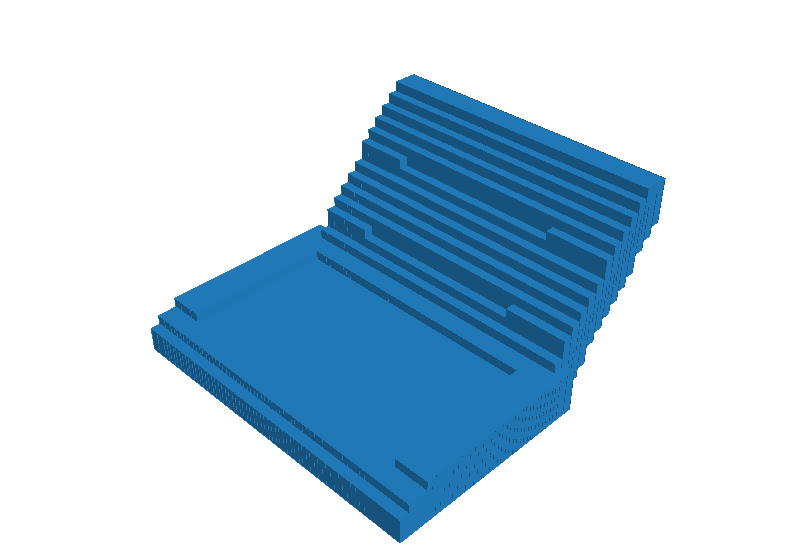} \\


\includegraphics[width=0.14\textwidth]{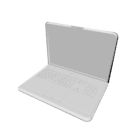} &
\includegraphics[width=0.20\textwidth]{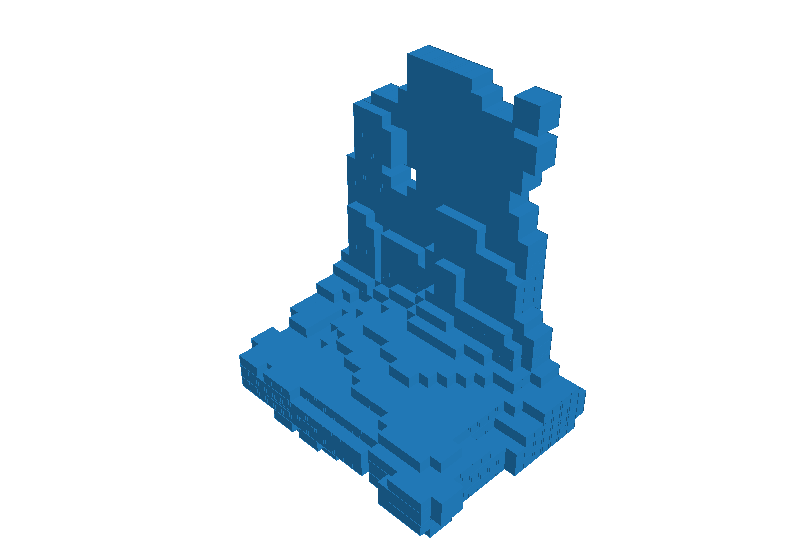} &
\includegraphics[width=0.20\textwidth]{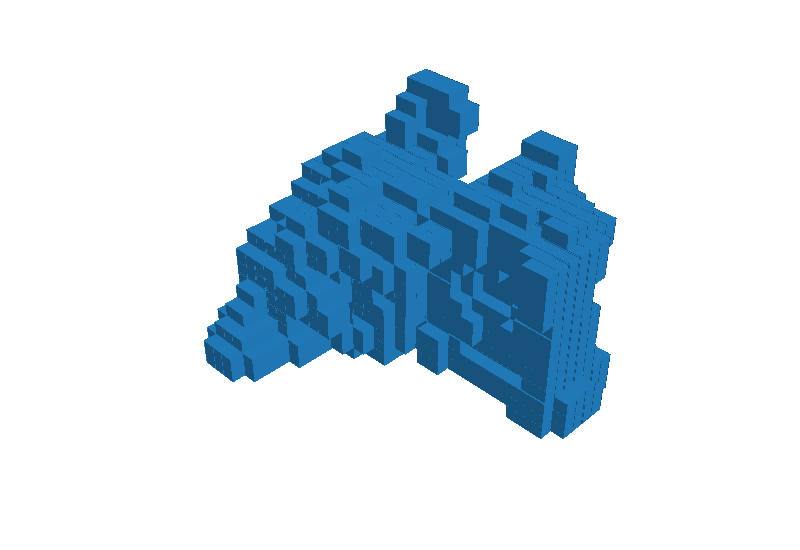} &
\includegraphics[width=0.20\textwidth]{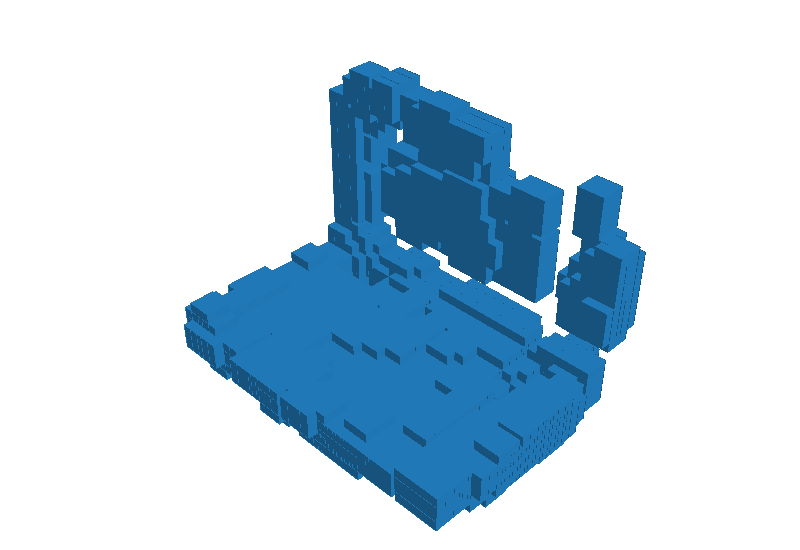} &
\includegraphics[width=0.20\textwidth]{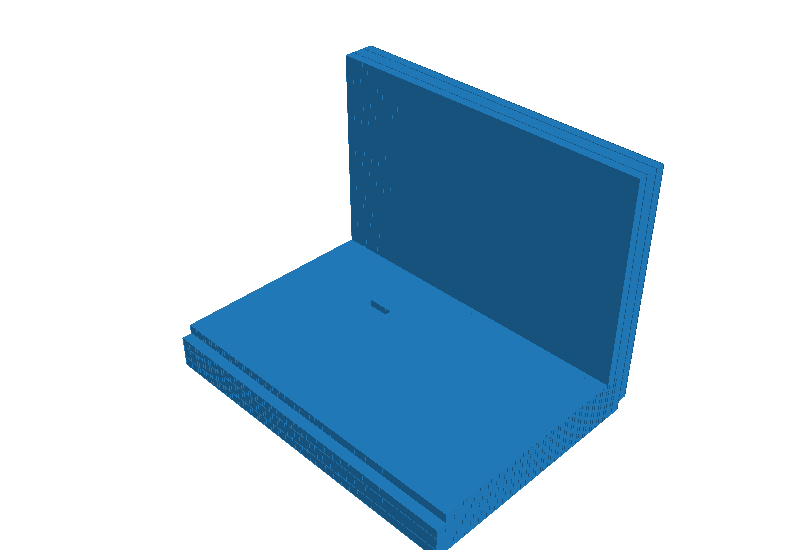} \\

\includegraphics[width=0.14\textwidth]{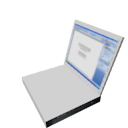} &
\includegraphics[width=0.20\textwidth]{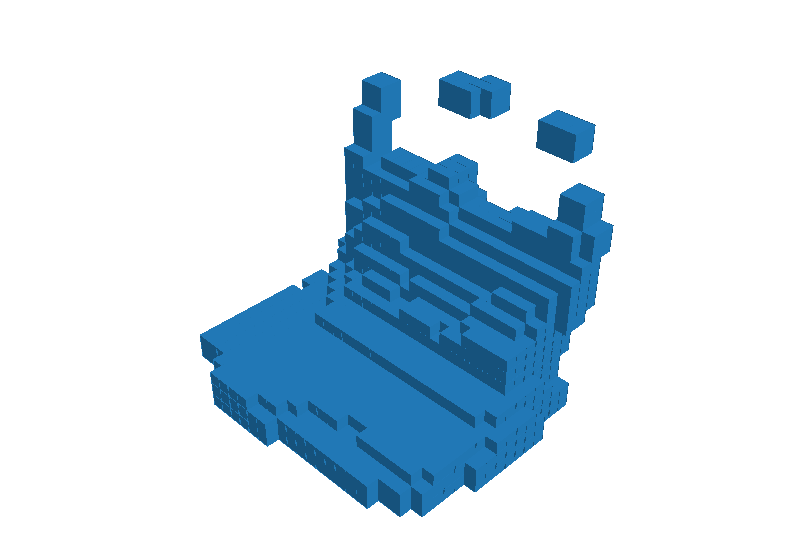} &
\includegraphics[width=0.20\textwidth]{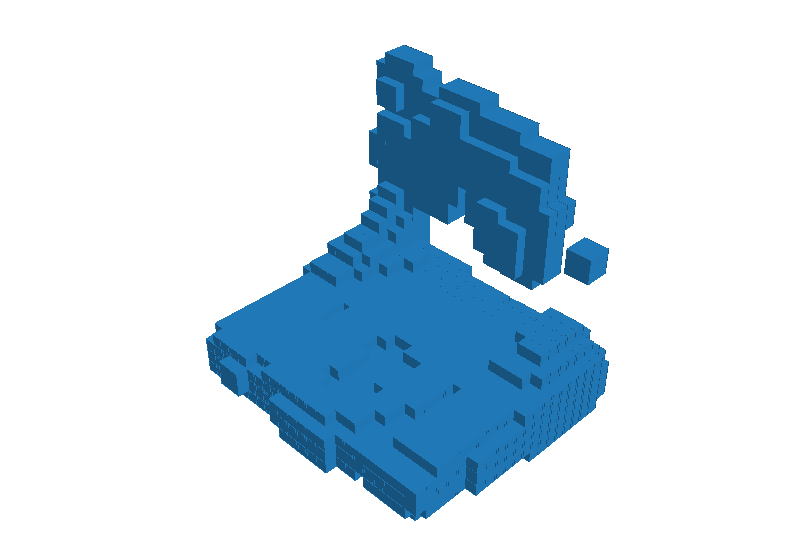} &
\includegraphics[width=0.20\textwidth]{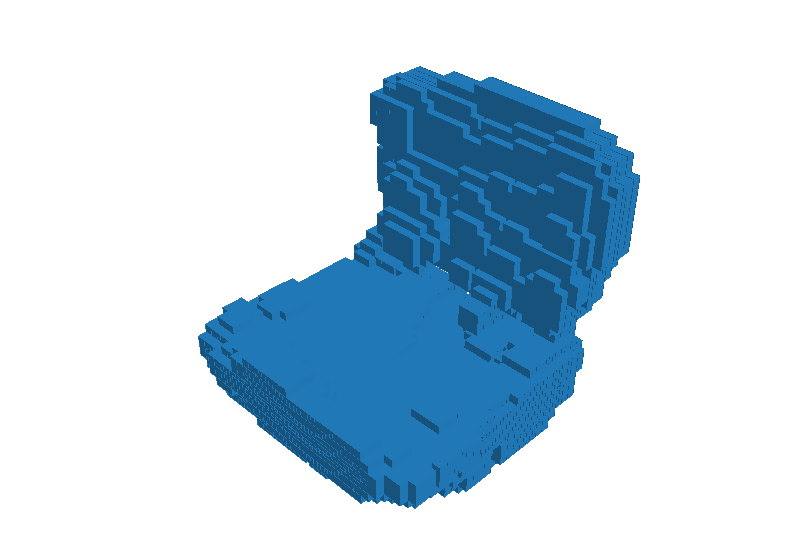} &
\includegraphics[width=0.20\textwidth]{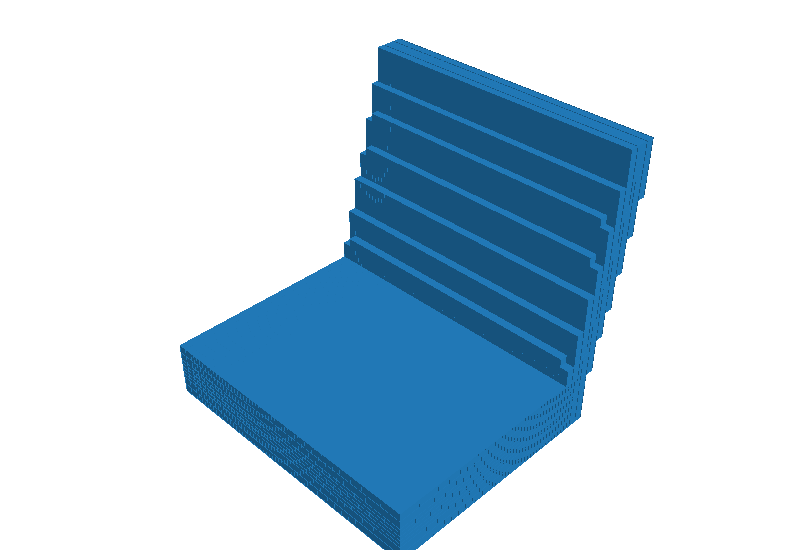} \\

\includegraphics[width=0.12\textwidth]{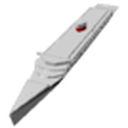} &
\includegraphics[width=0.20\textwidth]{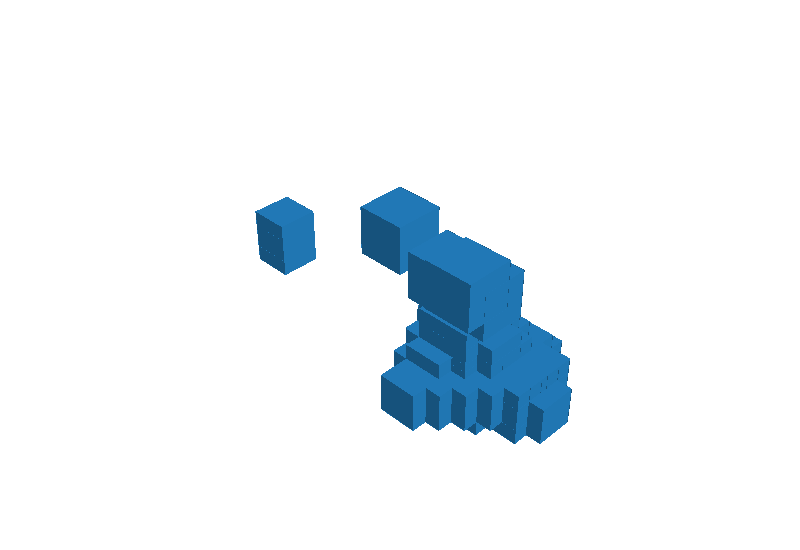} &
\includegraphics[width=0.20\textwidth]{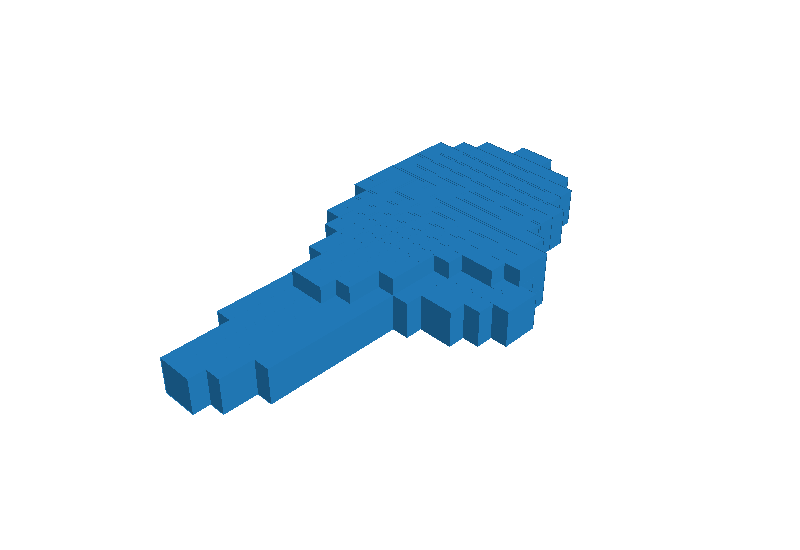} &
\includegraphics[width=0.20\textwidth]{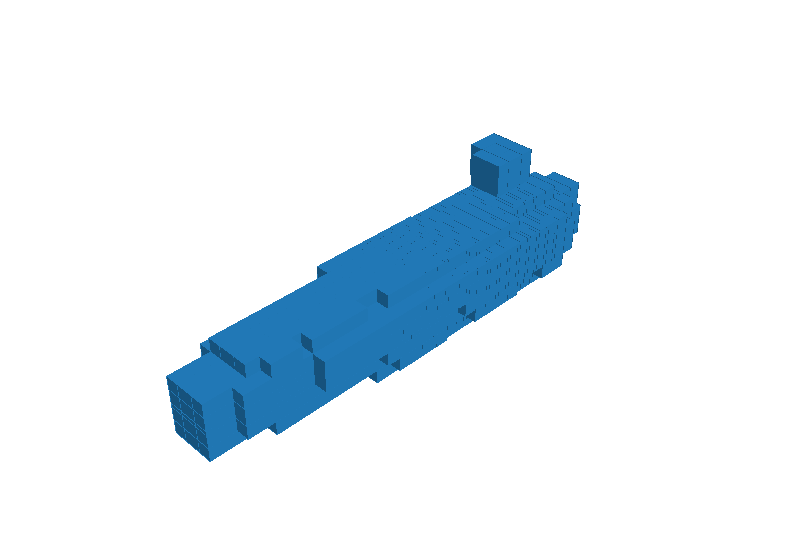} &
\includegraphics[width=0.20\textwidth]{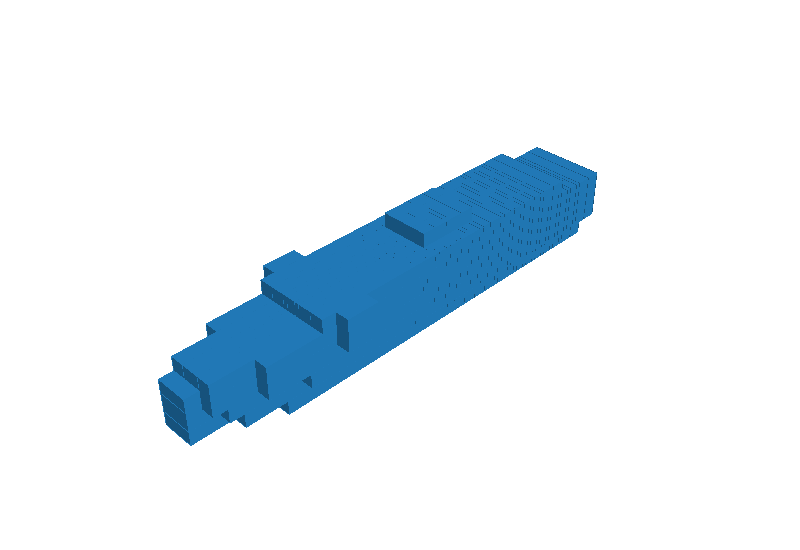} \\

\includegraphics[width=0.12\textwidth]{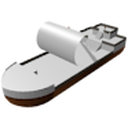} &
\includegraphics[width=0.20\textwidth]{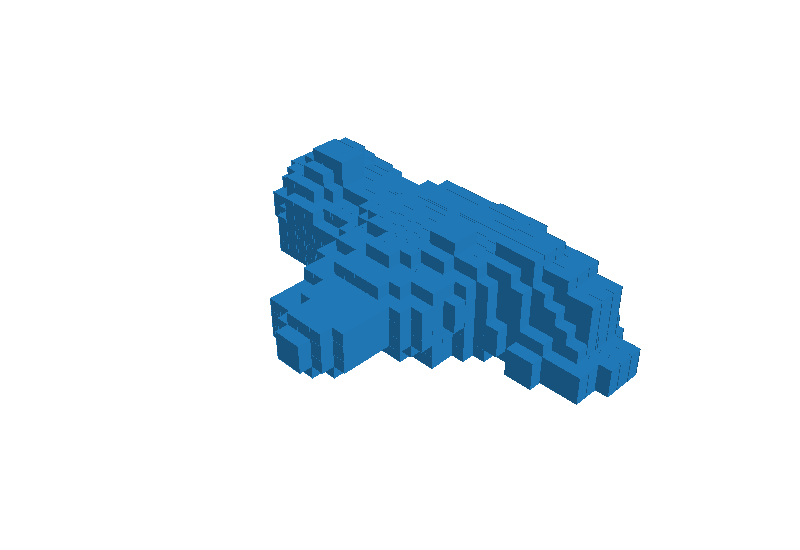} &
\includegraphics[width=0.20\textwidth]{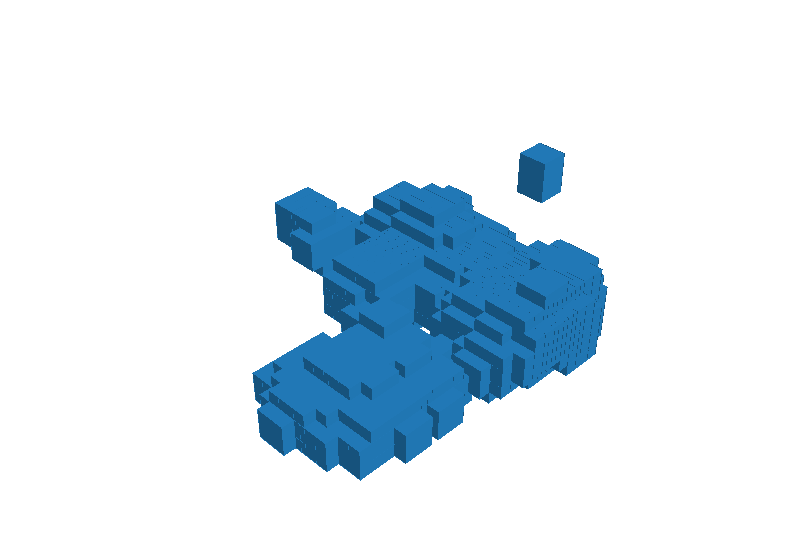} &
\includegraphics[width=0.20\textwidth]{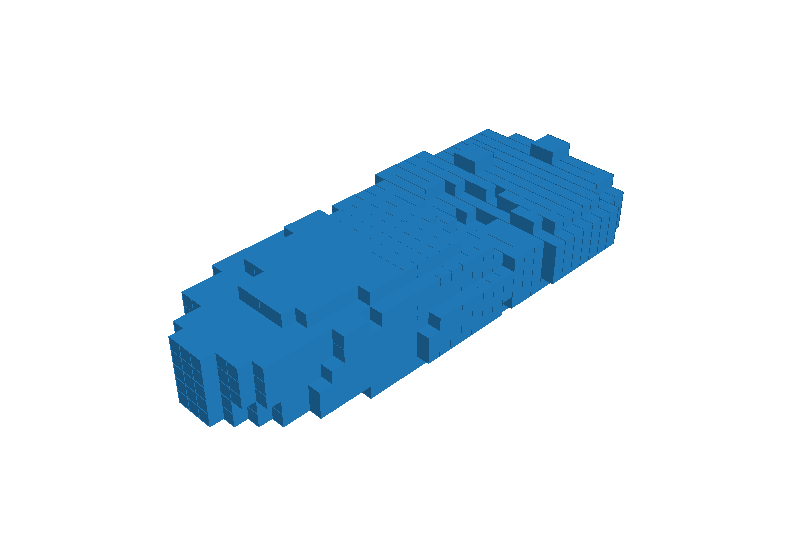} &
\includegraphics[width=0.20\textwidth]{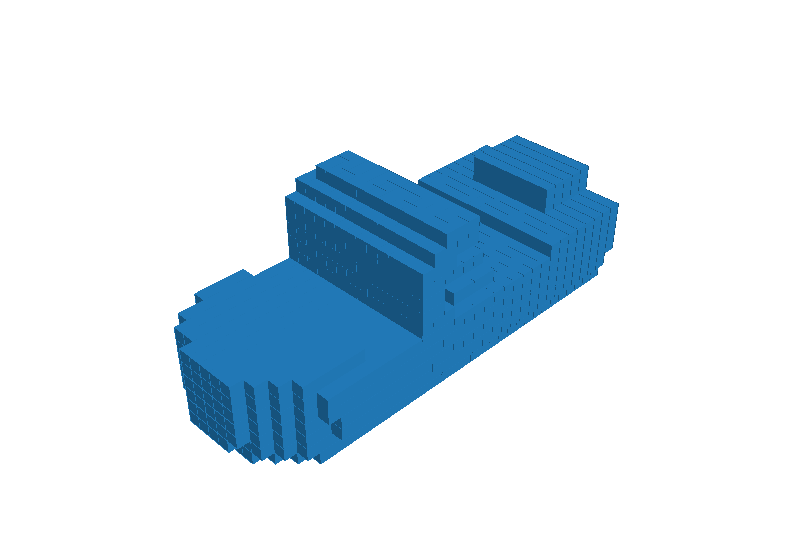} \\



\includegraphics[width=0.12\textwidth]{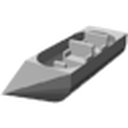} &
\includegraphics[width=0.20\textwidth]{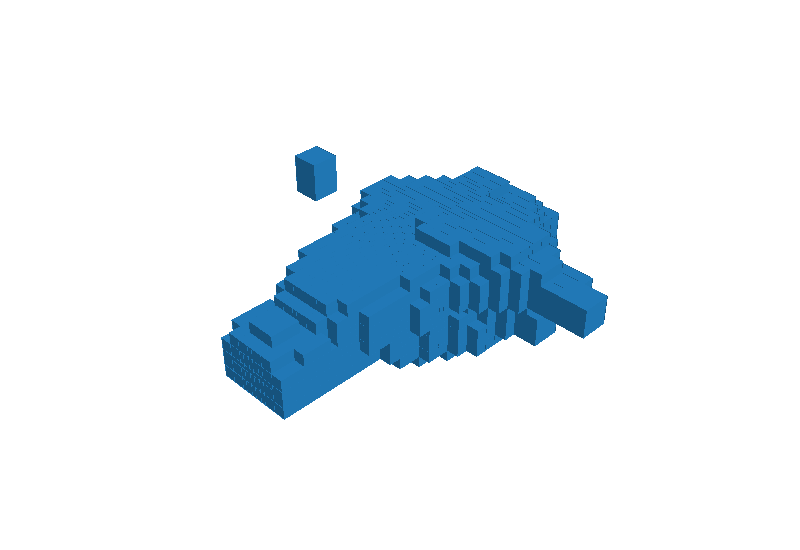} &
\includegraphics[width=0.20\textwidth]{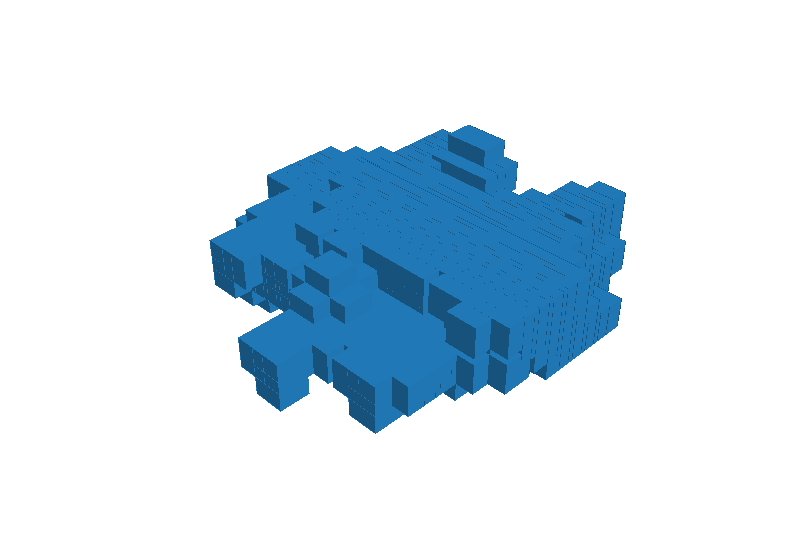} &
\includegraphics[width=0.20\textwidth]{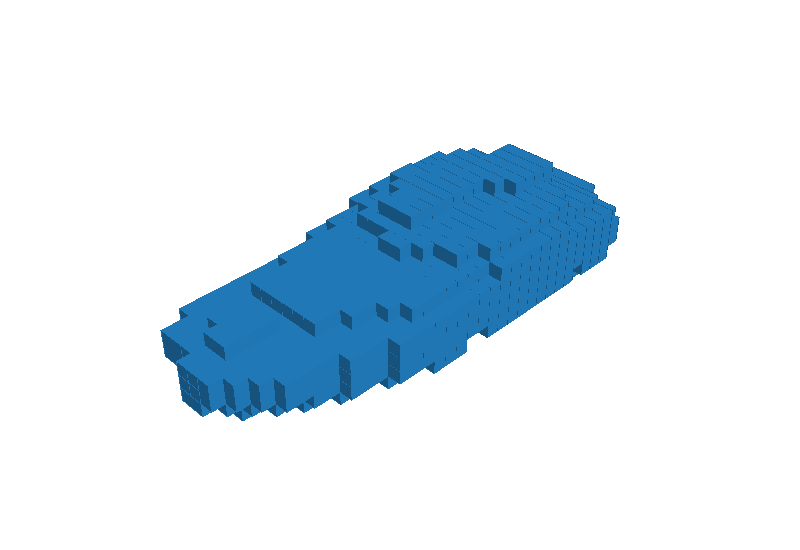} &
\includegraphics[width=0.20\textwidth]{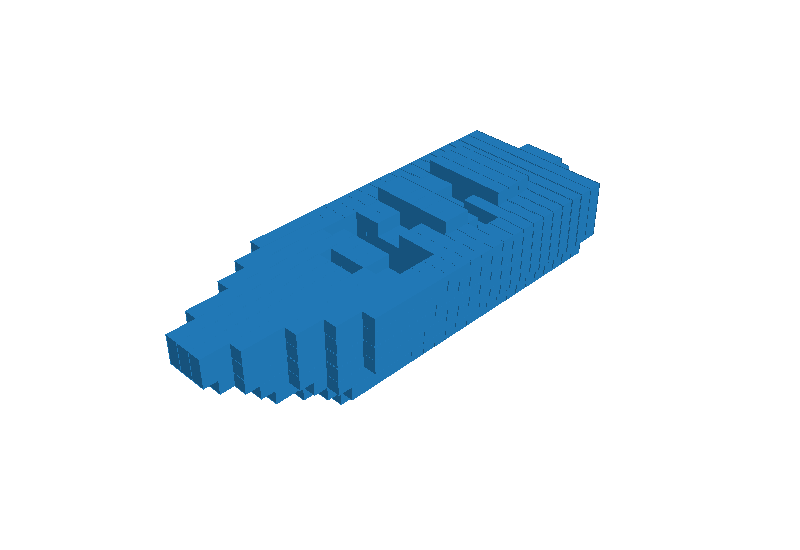} \\

\hline
\end{tabular}
\caption{Qualitative examples of shape inference obtained by Zero-Shot baseline, Walalce and proposed CGCE approach. We have used 2D views from random angles, but for visualization purposes the views are aligned to the same angle.}
\label{fig:1}
\end{figure}

\end{document}